\documentclass[journal]{IEEEtran}
\DeclareMathAlphabet{\mathcal}{OMS}{cmsy}{m}{n}
\usepackage{amsmath,amsfonts}
\usepackage{algorithmic}
\usepackage{algorithm}
\usepackage{amsthm}  
\usepackage{threeparttable}
\usepackage{threeparttable}
\usepackage{siunitx}
\usepackage{array}
\usepackage[caption=false,font=normalsize,labelfont=sf,textfont=sf]{subfig}
\usepackage{textcomp}
\usepackage{stfloats}
\usepackage{url}
\usepackage{graphicx}        
\usepackage{float}           
\usepackage{verbatim}
\usepackage[T1]{fontenc} 
\usepackage{mathrsfs} 
\usepackage{graphicx}
\graphicspath{{pic/}}
\usepackage{cite}
\usepackage{amssymb,amsthm}
\usepackage{multirow}
\usepackage[hidelinks]{hyperref}
\usepackage{booktabs,multirow}
\usepackage{booktabs,multirow}
\usepackage{booktabs,multirow}
\usepackage{float}
\usepackage{array}
\usepackage{subfig}
\usepackage[export]{adjustbox}  

\usepackage{caption}

\usepackage[utf8]{inputenc}
\usepackage{newunicodechar}
\newunicodechar{≈}{\ensuremath{\approx}}

	\makeatletter
	\newtheorem{lemma}{Lemma}
	\newtheorem{theorem}{Theorem}
	\newtheorem{definition}{Definition}
	\newtheorem{proposition}{Proposition}
	\newtheorem{assumption}{Assumption}
	\newtheorem{corollary}{Corollary}
	
	\newcommand{\Rmnum}[1]{\expandafter\@slowromancap\romannumeral #1@}
	\makeatother
	
	\def\BibTeX{{\rm B\kern-.05em{\sc i\kern-.025em b}\kern-.08em
			T\kern-.1667em\lower.7ex\hbox{E}\kern-.125emX}}
	\usepackage{balance}

	\title{Risk-Consistent Multiclass Learning from Random Label-Subset Membership Queries}
	
	\author{Jiaxu Su, Junpeng Li, \emph{Member}, \emph{IEEE}, Changchun Hua, \emph{Fellow}, \emph{IEEE} and Yana Yang, \emph{Member}, \emph{IEEE}%
		\thanks{J. Su, J. Li, C. Hua, and Y. Yang are with the Engineering Research Center of the Ministry of Education for Intelligent Control System and Intelligent Equipment, Yanshan University, Qinhuangdao, China 
			(e-mail: sjx67912@gmail.com; jpl@ysu.edu.cn; cch@ysu.edu.cn; yyn@ysu.edu.cn).}}

\begin{document}
		\maketitle
		
	\begin{abstract}
		Obtaining accurate class labels is often costly or unreliable, and may also be limited by privacy or other practical conditions. Compared with asking an annotator to provide the exact class, it is often easier to ask whether the true label belongs to a certain label subset. This \emph{query-response} form defines a distinct weak-supervision mechanism: weak supervision information is generated through feedback on a label subset. Although weakly supervised learning has studied many learning frameworks, most existing work starts from established weak-label objects. A systematic characterization is still lacking for weakly supervised learning generated directly by such \emph{query-response} observations. This paper proposes a multiclass learning framework under \emph{random label-subset queries}. We model the data-generating distribution of query-response observations and derive an unbiased estimator of the target risk under the empirical risk minimization (ERM) framework. To address negative empirical risk and the associated overfitting problem, we introduce corrected risk estimators based on non-negative and absolute-value corrections. Theoretical analysis establishes a conditional generalization and excess-risk bound for the unbiased estimator, and a bias-and-consistency result for the corrected risk estimator. Experiments under the matched random-query mechanism demonstrate the feasibility of direct query-response learning and the stabilization effect of risk correction.
	\end{abstract}
	\begin{IEEEkeywords}
		Query-response weak supervision, label-subset membership queries, risk rewriting, corrected risk estimators.
	\end{IEEEkeywords}
	
	\section{Introduction}
	
	Large-scale supervised learning typically relies on training examples paired with exact class labels.
	In many practical applications, however, obtaining such labels is expensive, unreliable, or constrained by privacy, expertise, or annotation interfaces.
	This has motivated a broad family of weakly supervised learning problems, where the learner aims to build a predictive model from supervision that is incomplete, inexact, inaccurate, indirect, or aggregated~\cite{zhou2018brief}.
	Representative examples include positive-unlabeled learning~\cite{elkan2008learning,kiryo2017positive}, complementary-label learning~\cite{ishida2017learning,ishida2019complementary,feng2020learning,chou2020unbiased}, partial-label learning~\cite{cour2011learning,lv2020progressive,feng2020provably}, learning from label proportions~\cite{quadrianto2009estimating,7549044,9718021}, aggregate-observation learning~\cite{zhang2020learning}, and pairwise or similarity-based supervision~\cite{hsu2019multi,cao2021learning}.
	Although these settings differ in the form of observed supervision, they often share the same statistical goal: to recover or approximate the ordinary supervised prediction target from weak observations.
	
	A common approach in weakly supervised learning is to begin with a predefined weak-label object.
	For instance, positive-unlabeled learning observes positive and unlabeled examples~\cite{elkan2008learning,kiryo2017positive}; complementary-label learning observes labels that an instance does not belong to~\cite{ishida2017learning,ishida2019complementary}; partial-label learning observes candidate sets that contain the true label~\cite{cour2011learning,lv2020progressive,feng2020provably}; and learning from label proportions observes aggregate class proportions for bags of instances~\cite{quadrianto2009estimating,7549044,9718021}.
	Recent general frameworks further study how different weak-supervision regimes can be understood through risk rewriting, latent-variable modeling, or unified weak-supervision representations~\cite{chen2024general,chiang2025unified}.
	These developments have provided important foundations for learning from weak annotations.
	Nevertheless, in many real data-collection scenarios, weak supervision is not simply handed to the learner as a fixed weak-label object.
	Instead, it is generated by an interaction or querying process.
	
	This paper studies such a query-response form of weak supervision for multiclass classification.
	Let $\mathcal{Y}=\{1,\ldots,k\}$ be the label space.
	For each latent example $(x,y)$, the learner does not observe the true label $y$.
	Instead, a label subset $L\subset\mathcal{Y}$ of fixed size $m$ is queried, and the learner observes only a binary membership response
	\begin{equation*}
		s=\mathbf{1}\{y\in L\}.
	\end{equation*}
	Thus, the training data consist of triples $(x,L,s)$, where $L$ is the queried label subset and $s\in\{0,1\}$ indicates whether the true label belongs to that subset.
	We refer to this observation protocol as a random label-subset membership query.
	Intuitively, rather than asking an annotator to reveal the exact class, the learner asks a broader yes/no question: ``Does the true class belong to this subset of labels?''
	Such questions may be easier to answer, less privacy-sensitive, or more compatible with restricted annotation settings.
	
	At the level of individual observations, this supervision may appear reducible to existing weak-label forms.
	If $s=1$, then the queried subset $L$ contains the true label and resembles a partial-label candidate set~\cite{cour2011learning,lv2020progressive,feng2020provably}.
	If $s=0$, then all labels in $L$ are incorrect and the subset resembles multiple complementary labels~\cite{ishida2017learning,ishida2019complementary,feng2020learning}.
	However, this observation-level conversion does not preserve the full data-generating mechanism.
	In our setting, both response groups are induced by the same randomized subset-query process and are coupled through the same latent supervised distribution $p(x,y)$.
	The query mechanism determines not only which weak-label object is observed for an individual sample, but also the response-group proportions and the response-conditioned distributions $p(x,L\mid s=1)$ and $p(x,L\mid s=0)$.
	Therefore, the central question is not merely how to convert query-response observations into partial-label or complementary-label data, but whether the query-response mechanism itself admits an exact recovery of the ordinary supervised risk.
	
		\begin{figure*}[!t]
		\centering
		\includegraphics[width=0.95\textwidth]{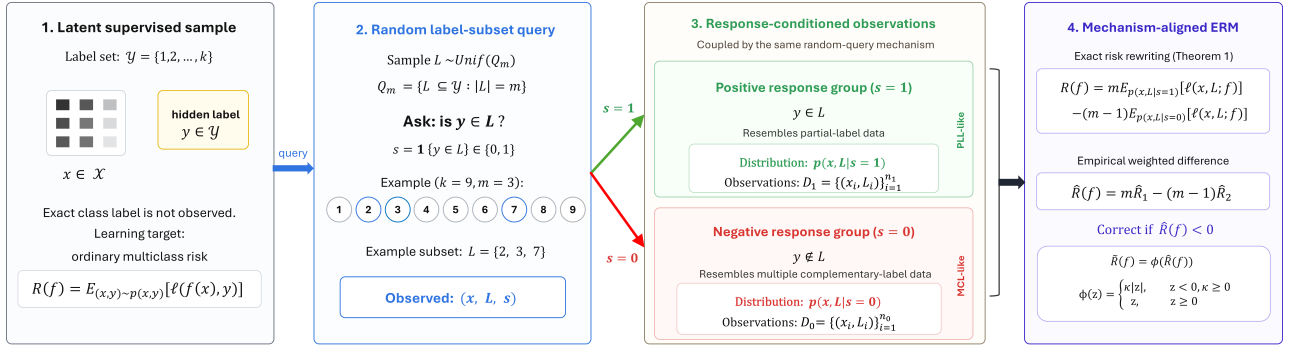}
		\caption{Mechanism-aware risk recovery from random label-subset membership queries. For each latent supervised sample $(x,y)$, the learner does not observe the exact class label. Instead, a label subset $L\subset\mathcal{Y}$ of size $m$ is randomly queried, and only the binary membership response $s=\mathbf{1}\{y\in L\}$ is observed. Positive responses resemble partial-label observations, while negative responses resemble multiple complementary-label observations. However, the two response groups are coupled by the same random-query mechanism and should be modeled jointly. This mechanism induces the exact risk identity $R(f)=m\mathbb{E}[\bar{\ell}(X,L;f)\mid s=1]-(m-1)\mathbb{E}[\bar{\ell}(X,L;f)\mid s=0]$, leading to a mechanism-aligned empirical risk estimator and corrected objectives for stabilizing negative empirical risk.}
		\label{fig:query_response_risk_recovery}
	\end{figure*}
	
	We answer this question under a symmetric random-query model.
	We first characterize the observable distributions induced by the query mechanism.
	Conditioning on a positive response restricts the latent class to labels inside the queried subset, whereas conditioning on a negative response restricts the latent class to labels outside the queried subset.
	This yields explicit expressions for $p(x,L\mid s=1)$ and $p(x,L\mid s=0)$, which act as mechanism-specific bridges between the observable query-response data and the latent supervised distribution.
	Based on these response-conditioned distributions and elementary combinatorial identities, we derive an exact rewriting of the ordinary supervised multiclass risk.
	
	Specifically, for a classwise loss $\ell(f(x),y)$, let
	\begin{equation*}
		\bar{\ell}(x,L;f)=\frac{1}{m}\sum_{j\in L}\ell(f(x),j)
	\end{equation*}
	be the average loss over the queried subset.
	Under the symmetric random-query mechanism, the supervised target risk can be written as
	\begin{equation*}
		R(f)
		=
		m\,\mathbb{E}[\bar{\ell}(X,L;f)\mid s=1]
		-
		(m-1)\,\mathbb{E}[\bar{\ell}(X,L;f)\mid s=0].
	\end{equation*}
	This identity shows that the ordinary supervised risk can be recovered directly from the original query-response observations.
	The resulting empirical objective is therefore mechanism-aligned: it is constructed from the two response groups generated by the same subset-query process, rather than from a post-hoc conversion to an existing weak-label format.
	
	The empirical estimator induced by this identity is a weighted difference of two response-group sample means. This structure enables unbiased risk recovery, but it can also produce negative empirical risk in finite samples. This issue is related to negative-risk and overfitting phenomena observed in other unbiased weak-supervision risk estimators \cite{kiryo2017positive,ishida2019complementary,chou2020unbiased}, although here it arises from the response-group difference induced by label-subset membership queries.
	
	To address negative empirical risk, we introduce corrected risk estimators based on non-negative and absolute-value corrections. These corrections modify the empirical objective when the raw estimator enters a negative region, without redefining the population learning target.
	
	We evaluate the proposed framework under synthetic query mechanisms matched to the theory.
	Starting from fully labeled multiclass datasets, we hide the true labels from the learner, randomly generate queried label subsets, and reveal only the binary membership responses.
	This experimental design isolates the statistical question studied in the theory.
	The results show that direct learning from query-response supervision is feasible under the matched random-query mechanism.
	They also show that correction can substantially improve robustness when the raw weighted-difference estimator becomes unstable.
	We emphasize that these experiments are intended to validate the proposed mechanism-aligned formulation under its theoretical assumptions, rather than to claim universal superiority over all transformed weak-label baselines.

	The main contributions of this paper are summarized as follows:
	\begin{itemize}
		\item We formulate multiclass learning from random label-subset membership queries as a query-response weak-supervision problem. In this setting, supervision is generated by a subset-query mechanism rather than given as a predefined weak-label object.
		
		\item Under a symmetric random-query model, we characterize the induced response-conditioned distributions and derive an exact rewriting of the ordinary supervised multiclass risk. This identity leads to a direct unbiased risk estimator based on the original query-response observations.
		
		\item We analyze the statistical and optimization behavior of the resulting estimator. We establish conditional generalization and excess-risk bounds for the unbiased estimator, identify its weighted difference-of-means structure as the source of negative empirical risk, and establish a bias-and-consistency result for the corrected risk estimator.
		
		\item We evaluate the proposed framework under matched synthetic query mechanisms and compare it with transformed partial-label and multiple complementary-label baselines. The experiments demonstrate the feasibility of direct query-response learning and the stabilization effect of risk correction, while clarifying the scope and limitations of the proposed formulation.
	\end{itemize}
	
	The rest of this paper is organized as follows. Section II reviews the most relevant literature. Section III introduces the random label-subset query mechanism and derives the risk rewriting. Section IV analyzes the induced estimator, finite-sample instability, and corrected objectives. Section V reports the experimental results. The main proofs are provided in the appendix.
	
	\section{Related Work}
	
	\subsection{Weakly supervised learning and mechanism-aware risk recovery}
	
	Weakly supervised learning studies predictive learning when exact labels are unavailable, unreliable, or replaced by weaker forms of supervision \cite{zhou2018brief}. Representative settings include positive-unlabeled learning \cite{elkan2008learning,kiryo2017positive}, complementary-label learning \cite{ishida2017learning,ishida2019complementary,feng2020learning,chou2020unbiased}, partial-label learning \cite{cour2011learning,lv2020progressive,feng2020provably}, learning from label proportions \cite{quadrianto2009estimating,7549044,9718021}, aggregate-observation learning \cite{zhang2020learning}, and similarity-based supervision \cite{hsu2019multi,cao2021learning}. Although the observable supervision differs across these settings, the target is often the ordinary supervised prediction problem.
	
	A common theoretical route is risk recovery or risk rewriting: one derives an objective computable from weak observations whose population counterpart equals, or is consistent with, the desired supervised risk. This viewpoint appears in PU learning \cite{elkan2008learning,kiryo2017positive}, complementary-label learning \cite{ishida2017learning,ishida2019complementary}, multiple complementary-label learning \cite{feng2020learning}, and risk-consistent partial-label learning \cite{feng2020provably}. More general perspectives have also studied weak supervision through unified representations or unified risk-analysis formulations \cite{chen2024general,chiang2025unified}. Our work follows this mechanism-aware line, but considers a different observation process in which a queried label subset and a binary membership response are generated by the same random query mechanism.
	
	\subsection{Positive-unlabeled learning and non-negative risk correction}
	
	Positive-unlabeled learning studies binary classification from labeled positive examples and unlabeled examples \cite{elkan2008learning}. Kiryo et al. \cite{kiryo2017positive} showed that unbiased PU risk estimators may become negative and cause severe overfitting when optimized with flexible models, and proposed a non-negative risk estimator to improve empirical stability.
	
	Our problem is not a PU-learning problem, since the target is multiclass classification and the weak supervision is generated by label-subset membership queries rather than positive selection and unlabeled sampling. Nevertheless, PU learning is methodologically related: both settings recover a supervised target risk from indirect observations, and both show that unbiased risk estimators can be statistically correct yet empirically unstable. In our setting, negative empirical risk arises from a weighted difference of two response-conditioned sample means induced by the subset-query mechanism.
	
	\subsection{Complementary-label, multiple complementary-label, and partial-label learning}
	
	Complementary-label learning assumes that each instance is annotated with a label that it does not belong to \cite{ishida2017learning}. This line has been extended to arbitrary losses and models \cite{ishida2019complementary}, biased complementary labels \cite{yu2018learning}, and multiple complementary labels \cite{feng2020learning}. Partial-label learning assumes that each example is associated with a candidate-label set containing the true label \cite{cour2011learning}. Recent methods such as PRODEN perform candidate-label disambiguation in deep models \cite{lv2020progressive}, while risk-consistent partial-label learning makes the candidate-label generation mechanism explicit \cite{feng2020provably}. The empirical instability of unbiased estimators in complementary-label learning has also been analyzed in \cite{chou2020unbiased}.
	
	Our query-response setting is related to both directions. When \(s=0\), all labels in the queried subset \(L\) are incorrect, so the observation resembles multiple complementary labels. When \(s=1\), the queried subset contains the true label, so it resembles a partial-label candidate set. However, these two cases are generated by the same random subset-query mechanism and are coupled through the same latent distribution \(p(x,y)\). Therefore, the proposed risk identity uses both response-conditioned laws jointly, rather than treating the positive and negative response groups as separate weak-label datasets.
	
	\subsection{Learning from aggregate, pairwise, and indirect supervision}
	
	Several weak-supervision settings replace instance-level labels with aggregate or relational information. Learning from label proportions observes class proportions over bags of instances \cite{quadrianto2009estimating,7549044,9718021}, aggregate-observation learning studies broader group-level supervision \cite{zhang2020learning}, and pairwise or similarity-based learning replaces class labels with relational information \cite{hsu2019multi,cao2021learning}. Recent studies further explore weak supervision from pairwise confidence comparisons with unlabeled data and from \(M\)-tuple similarity-confidence data \cite{li2024learning,li2025binary}. These settings show that ordinary predictors can sometimes be learned from indirect observations when the observation mechanism is sufficiently structured.
	
	The observation unit in our setting is different. We do not observe bag proportions, pairwise similarities, or similarity-confidence scores. Each observation consists of a single instance, a queried label subset, and a binary membership response. The technical challenge is therefore to characterize how this label-subset membership query mechanism transforms the latent distribution \(p(x,y)\) into observable response-conditioned laws.
	
	\subsection{Query-based supervision and partial feedback}
	
	The proposed setting is also related to query-based supervision and active learning, where a learner may select informative queries to reduce labeling cost \cite{settles2009active}. Recent query-based annotation designs have explored alternatives to requesting full labels, such as local multi-class information in semantic segmentation or other structured annotation feedback \cite{hwang2023active}. Recent application-oriented studies have also explored annotation-budget reduction in large-scale sleep staging under active-learning settings \cite{liu2024activesleeplearner}. These works mainly emphasize query selection or annotation efficiency.
	
	Our focus is different. We do not study adaptive query selection. Instead, we fix a symmetric random-query mechanism and analyze the statistical learning problem induced by the resulting query-response observations. Extending the analysis to adaptive, non-uniform, or noisy label-subset membership queries is left for future work.
	
	\subsection{Position of the present work}
	
	The present work sits between weak-label risk recovery and query-response supervision. At the observation level, positive responses resemble partial-label data and negative responses resemble multiple complementary-label data. However, the two response groups are not generated independently. They are coupled by the same random label-subset membership query mechanism, and this coupling determines the response-group proportions, the response-conditioned laws, and the exact risk identity. Thus, our contribution is not merely a conversion of query-response observations into existing weak-label objects, but a mechanism-aware risk-recovery analysis for the original query-response data.

	\section{RANDOM LABEL-SUBSET QUERIES AND MECHANISM-INDUCED RISK RECOVERY}\label{sec:problem}
		
	This section formalizes the random label-subset membership query mechanism and derives the supervised-risk identity induced by this mechanism. The statistical target is the ordinary multiclass supervised risk, whereas the learner only observes query-response triples $(x,L,s)$, where $L$ is a queried label subset and $s$ is a binary membership response. Our goal is not to convert these observations into pre-existing weak-label objects, but to characterize the observable law generated by the query process and to recover the supervised target directly from the resulting response-conditioned distributions.
		
		\subsection{Problem Setting}
		
		Let $\mathcal{Y}:=\{1,\ldots,k\}$ denote the label space, and let $\mathcal{X}$ denote the feature space. We assume that latent samples $(x,y)$ are drawn i.i.d. from an unknown joint distribution with density $p(x,y)$ on $\mathcal{X}\times\mathcal{Y}$, and we write $p(x)$ for the corresponding marginal density. For a fixed $m\in\{1,\ldots,k-1\}$, define
		\[
		\mathcal{Q}_m:=\binom{\mathcal{Y}}{m},
		\]
		that is, the family of all label subsets of $\mathcal{Y}$ with cardinality exactly $m$. Thus, each sample-specific query subset $L$ takes values in $\mathcal{Q}_m$.
		
		We consider prediction functions $f:\mathcal{X}\to\mathbb{R}^k$ and a classwise loss $\ell:\mathbb{R}^k\times\mathcal{Y}\to\mathbb{R}$. The ordinary supervised target risk is
		\begin{equation}\label{eq:supervised}
			R(f):=E_{(x,y)\sim p(x,y)}[\ell(f(x),y)].
		\end{equation}
		
		The learner does not observe the exact label $y$ directly. Instead, it observes triples $(x,L,s)$, where $L\in\mathcal{Q}_m$ is a queried label subset and
		\[
		s=\mathbf{1}\{y\in L\}\in\{0,1\}
		\]
		is the corresponding membership response. Thus, the observable supervision is query-response based, while the learning target remains the supervised risk $R(f)$.
		
		\subsection{Data Generation Process}
		
		We now specify the observation mechanism. Since the true label is latent, the observable information is entirely summarized by whether the queried subset contains the true class. This makes the two response-conditioned laws
		\[
		p(x,L\mid s=1)
		\qquad\text{and}\qquad
		p(x,L\mid s=0)
		\]
		the natural basic objects of the analysis.
		
		\begin{assumption}\label{assumption1}
			For each latent sample $(x,y)$, the queried subset $L$ is drawn uniformly from $\mathcal{Q}_m$, that is,
			\begin{equation}
				P(L=L'\mid x,y)=
				\begin{cases}
					\binom{k}{m}^{-1}, & L'\in\mathcal{Q}_m,\\
					0, & \text{otherwise},
				\end{cases}
				\label{eq:query-mech}
			\end{equation}
			and the binary response is generated deterministically by
			\begin{equation}
				s:=\mathbf{1}\{y\in L\}.
				\label{eq:response-mech}
			\end{equation}
			The training data $\mathcal{D}=\{(x_i,L_i,s_i)\}_{i=1}^n$ are i.i.d. samples from the induced observable law of $(x,L,s)$.
		\end{assumption}
		
		Under this mechanism, the response-group proportions depend only on $(k,m)$.
		
		\begin{lemma}
			The following equality holds:
			\[
			P(s=\sigma)=
			\begin{cases}
				\frac{k-m}{k}, & \sigma=0,\\[0.4em]
				\frac{m}{k}, & \sigma=1.
			\end{cases}
			\]
		\end{lemma}
		
		Lemma~1 shows that the observable sample is split into two response groups with fixed population proportions. The next proposition identifies the corresponding conditional distributions of $(x,L)$.
		
		\begin{proposition}[Response-conditioned distributions]
			Under Assumption~1, for every $L\in\mathcal{Q}_m$,
			\begin{equation}
				\begin{aligned}
					p(x,L\mid s=1)&=\frac{1}{\binom{k-1}{m-1}}\sum_{y\in L}p(x,y),\\
					p(x,L\mid s=0)&=\frac{1}{\binom{k-1}{m}}\sum_{y\notin L}p(x,y).
				\end{aligned}
				\label{eq:conditional-law}
			\end{equation}
			Moreover, $p(x,L\mid s=\sigma)=0$ for all $L\notin\mathcal{Q}_m$ and $\sigma\in\{0,1\}$.
		\end{proposition}
		
		Proposition~1 is the first mechanism-specific bridge from the observable query-response law back to the latent supervised law. Conditioning on $s=1$ restricts the latent class to labels inside $L$, whereas conditioning on $s=0$ restricts it to labels outside $L$.
		
		\subsection{Risk Rewriting and Empirical Objective}
		
		Using Proposition~1 together with elementary combinatorial identities, one can express each joint quantity $p(x,y)$ in terms of the observable marginal $p(x)$ and either response-conditioned law.
		
		\begin{lemma}
			The following equalities hold:
			\[
			p(x,y)=\frac{k-1}{k-m}\sum_{L\in\mathcal{Q}^{\mathrm{in}}_y}p(x,L\mid s=1)-\frac{m-1}{k-m}p(x),
			\]
			\[
			p(x,y)=\frac{k-1}{m}\sum_{L\in\mathcal{Q}^{\mathrm{out}}_y}p(x,L\mid s=0)-\frac{k-m-1}{m}p(x),
			\]
			where
			\[
			\mathcal{Q}^{\mathrm{in}}_y:=\{L\in\mathcal{Q}_m:y\in L\},
			\qquad
			\mathcal{Q}^{\mathrm{out}}_y:=\{L\in\mathcal{Q}_m:y\notin L\}.
			\]
		\end{lemma}
		
		Either expression in Lemma~2 is sufficient for the derivation below; we use the $s=1$ representation for convenience.
		
		\begin{theorem}\label{theorem1}
			The supervised target risk in \eqref{eq:supervised} can be written as
			\begin{equation}
				\begin{aligned}
					R(f)=&m\,E_{(x,L)\sim p(x,L\mid s=1)}[\bar{\ell}(x,L;f)]\\&-(m-1)\,E_{(x,L)\sim p(x,L\mid s=0)}[\bar{\ell}(x,L;f)],
				\end{aligned}
				\label{eq:risk-identity}
			\end{equation}
			where
			\[
			\bar{\ell}(x,L;f):=\frac{1}{m}\sum_{j\in L}\ell(f(x),j).
			\]
		\end{theorem}
		
		Eq~\eqref{eq:risk-identity} is the central identification result of this section. It shows that the ordinary supervised risk can be recovered directly from the original query-response observations, without first recasting them as another weak-label format.
		The identity in Theorem~\ref{theorem1} also clarifies why the proposed formulation is not merely a post-hoc combination of partial-label and multiple complementary-label learning. A positive response makes L resemble a candidate-label set, whereas a negative response makes L resemble a set of complementary labels. However, the coefficients in Eq~\eqref{eq:risk-identity} and the two response-conditioned expectations are determined jointly by the same randomized query mechanism. Thus, the risk recovery relies on the coupling between the two response groups induced by $p(x,L,s)$, rather than on treating the two groups as independent weak-label datasets.

		Given i.i.d. data $\mathcal{D}=\{(x_i,L_i,s_i)\}_{i=1}^n$, Eq.~\eqref{eq:risk-identity} induces the empirical estimator
		\begin{equation}
			\widehat{R}(f)=\frac{m}{n_1}\sum_{i:s_i=1}\bar{\ell}_i-\frac{m-1}{n_0}\sum_{j:s_j=0}\bar{\ell}_j,
			\label{eq:empirical-risk}
		\end{equation}
		where $\bar{\ell}_i:=\bar{\ell}(x_i,L_i;f)$, $n_1:=\sum_{i=1}^n\mathbf{1}\{s_i=1\}$, and $n_0:=\sum_{i=1}^n\mathbf{1}\{s_i=0\}$. The estimator is defined on the event $\{n_1\ge 1,\ n_0\ge 1\}$.
		
		The form of \eqref{eq:empirical-risk} is important. The direct empirical objective is a weighted difference of two response-group sample means. This estimator form will later explain why finite-sample instability is structural rather than accidental.
		
		\subsection{Conditional Generalization and Excess-Risk Bound}
		
		We next quantify the estimation error of the groupwise empirical objective in \eqref{eq:empirical-risk}. The result is a standard ERM-type uniform-deviation bound specialized to the present two-group estimator. Its main purpose is to make explicit how the learning guarantee depends on the effective sample sizes of the two observable response groups.
		
		\begin{definition}[Rademacher complexity {\cite{mohri2018foundations}}]
			Let $Z_1,\ldots,Z_n$ be i.i.d. samples from a distribution $\mu$ on a measurable space $\mathcal{Z}$. Let $\mathcal{G}$ be a class of measurable functions $g:\mathcal{Z}\to\mathbb{R}$. The expected Rademacher complexity of $\mathcal{G}$ is
			\[
			\mathfrak{R}_n(\mathcal{G})
			:=
			E_{Z_{1:n}\sim\mu}\,
			E_{\sigma}
			\left[
			\sup_{g\in\mathcal{G}}
			\frac{1}{n}\sum_{i=1}^n \sigma_i g(Z_i)
			\right],
			\]
			where $\sigma_1,\ldots,\sigma_n$ are i.i.d. Rademacher variables taking values in $\{-1,+1\}$ uniformly. In Theorem~2, this scalar notion is applied to the induced loss class generated by the score-vector hypothesis class.
		\end{definition}
		
		Assume that $\mathcal{F}$ is a hypothesis class of score-vector functions $f:\mathcal{X}\to\mathbb{R}^k$ such that
		\[
		\sup_{f\in\mathcal{F},\,x\in\mathcal{X}}\|f(x)\|_{\infty}\le B_{\mathcal{F}}
		\]
		for some constant $B_{\mathcal{F}}>0$. Assume also that the loss is a mapping $\ell:\mathbb{R}^k\times\mathcal{Y}\to\mathbb{R}$ such that, for all $j\in\mathcal{Y}$ and all $u\in[-B_{\mathcal{F}},B_{\mathcal{F}}]^k$,
		\[
		0\le \ell(u,j)\le C_{\ell},
		\]
		and that for each $j\in\mathcal{Y}$, the map $u\mapsto \ell(u,j)$ is $\rho$-Lipschitz with respect to $\|\cdot\|_{\infty}$ on $[-B_{\mathcal{F}},B_{\mathcal{F}}]^k$, i.e.,
		\[
		|\ell(u,j)-\ell(v,j)|\le \rho\|u-v\|_{\infty},
		\qquad
		\forall u,v\in[-B_{\mathcal{F}},B_{\mathcal{F}}]^k.
		\]
		For each $j\in\{1,\ldots,k\}$, define the coordinate function class
		\[
		\mathcal{F}_j:=\{x\mapsto f_j(x): f\in\mathcal{F}\}.
		\]
		Define the induced scalar loss class
		\[
		\mathcal{G}_{\mathcal{F}}
		:=
		\bigl\{(x,L)\mapsto \bar{\ell}(x,L;f)\;:\;f\in\mathcal{F}\bigr\}.
		\]
		The class $\mathcal{G}_{\mathcal{F}}$ is the scalar loss class induced by the score-vector hypothesis class $\mathcal{F}$; the coordinate classes $\mathcal{F}_j$ make explicit the componentwise structure underlying this multiclass setting. The standing $n^{-1/2}$ complexity-growth condition for $\mathcal{G}_{\mathcal{F}}$, together with its associated constant $C_R$, is stated separately as Lemma~D.3 in Appendix~D.
		
		For simplicity, we assume that the minimizers below exist. Let
		\[
		f^*\in\arg\min_{f\in\mathcal{F}}R(f),
		\qquad
		\hat{f}\in\arg\min_{f\in\mathcal{F}}\widehat{R}(f),
		\]
		where $\widehat{R}$ is the empirical objective in \eqref{eq:empirical-risk}.
		
		\begin{theorem}[Conditional uniform deviation and excess-risk bound]
			Fix any $\delta\in(0,1)$. Conditioning on the full indicator sequence $(s_1,\ldots,s_n)$, on the event $\{n_1\ge 1,\ n_0\ge 1\}$, with conditional probability at least $1-\delta$,
			\begin{align*}
				\sup_{f\in\mathcal{F}}
				\big|\widehat{R}(f)-R(f)\big|
				\le
				&m\left(
				\frac{2\rho C_R}{\sqrt{n_1}}
				+
				C_{\ell}\sqrt{\frac{\log(4/\delta)}{2n_1}}
				\right) \\
				&+
				(m-1)\left(
				\frac{2\rho C_R}{\sqrt{n_0}}
				+
				C_{\ell}\sqrt{\frac{\log(4/\delta)}{2n_0}}
				\right).
			\end{align*}
			Consequently, with the same conditional probability at least $1-\delta$,
			\begin{align*}
				R(\hat{f})-R(f^*)
				\le
				&\frac{4\rho m C_R}{\sqrt{n_1}}
				+2mC_{\ell}\sqrt{\frac{\log(4/\delta)}{2n_1}} \\
				&+\frac{4\rho (m-1) C_R}{\sqrt{n_0}}
				+2(m-1)C_{\ell}\sqrt{\frac{\log(4/\delta)}{2n_0}}.
			\end{align*}
		\end{theorem}
		
		\begin{corollary}[Unconditional form]
			Under Assumption~1, $n_1\sim \mathrm{Bin}(n,m/k)$ and $n_0=n-n_1$. Hence
			\[
			P(n_1=0)=\left(1-\frac{m}{k}\right)^n,
			\qquad
			P(n_0=0)=\left(\frac{m}{k}\right)^n.
			\]
			Therefore, with probability at least
			\[
			1-\delta-\left(1-\frac{m}{k}\right)^n-\left(\frac{m}{k}\right)^n,
			\]
			the event $\{n_1\ge 1,\ n_0\ge 1\}$ holds and the conclusions of Theorem~2 apply.
		\end{corollary}
		
		Theorem~2 is a standard uniform-deviation result specialized to the present two-group estimator. Its substantive content is the explicit dependence on the effective response-group sample sizes $n_1$ and $n_0$. When one observable group is small, the guarantee deteriorates accordingly. This dependence already suggests that population-level identifiability does not, by itself, imply benign finite-sample optimization behavior.
		
		\section{Negative Empirical Risk and Risk Correction}
		
		The risk identity in Section III induces a weighted difference of two response-conditioned sample means. This structure can produce negative empirical risk in finite samples. We therefore study scalar corrections that modify the empirical objective in negative regions while leaving the population target unchanged.
		
		\subsection{Negative Empirical Risk of the Raw Estimator}
		
		\begin{definition}[Groupwise empirical means]
			For a fixed $f\in\mathcal{F}$, define
			\begin{align*}
				&\widehat{R}_1(f):=\frac{1}{n_1}\sum_{i:s_i=1}\bar{\ell}(x_i,L_i;f), \\
				\qquad
				&\widehat{R}_0(f):=\frac{1}{n_0}\sum_{i:s_i=0}\bar{\ell}(x_i,L_i;f),
			\end{align*}
			so that the raw estimator can be written as
			\[
			\widehat{R}(f)=m\widehat{R}_1(f)-(m-1)\widehat{R}_0(f).
			\]
		\end{definition}
		
		Each groupwise mean is non-negative whenever the base loss is non-negative, but their weighted difference need not be. In particular,
		\[
		\frac{m-1}{n_0}\sum_{i:s_i=0}\bar{\ell}(x_i,L_i;f)
		>
		\frac{m}{n_1}\sum_{i:s_i=1}\bar{\ell}(x_i,L_i;f).
		\]
		Thus, negative empirical risk is a finite-sample effect produced by subtracting two separately estimated means. This behavior is consistent with prior observations in weak supervision and noisy-label learning, where population-correct estimators may still induce difficult optimization behavior in practice \cite{kiryo2017positive}. Related optimization-oriented analysis also appears in \cite{chou2020unbiased}.
		
		This observation motivates correcting the scalar empirical objective rather than modifying the underlying population target. We therefore consider
		\begin{equation}
			\widetilde{R}(f)=\phi(\widehat{R}(f)),
			\label{eq:corrected-risk}
		\end{equation}
		where
		\[
		\phi(z)=
		\begin{cases}
			z, & z\ge 0,\\
			\kappa|z|, & z<0,
		\end{cases}
		\qquad \kappa\ge 0.
		\]
		This family includes non-negative truncation when $\kappa=0$ and the absolute-value correction when $\kappa=1$. The correction acts only at the scalar empirical-risk level; it does not redefine the population learning target.
		
		We first study the corrected estimator at a fixed predictor.
		
		\begin{theorem}[Bias and consistency of the corrected risk estimator]\label{theorem3}
			Condition on the indicator sequence $s_1^n:=(s_1,\ldots,s_n)$ and assume $n_1\ge 1$ and $n_0\ge 1$. For any fixed predictor $f\in\mathcal{F}$, suppose that there exists a positive constant $\zeta_f$ such that $R(f)\ge \zeta_f$. Then the bias of the corrected risk estimator $\widetilde{R}(f)$ decreases at an exponential rate as $n_1,n_0\to\infty$:
			\[
			E[\widetilde{R}(f)]-R(f)
			\le
			(\kappa+1)(2m-1)C_\ell \, \Delta_f,
			\]
			where
			\[
			\Delta_f=
			\exp\!\left(
			-\frac{2\zeta_f^2}{
				C_\ell^2\left(\frac{m^2}{n_1}+\frac{(m-1)^2}{n_0}\right)}
			\right).
			\]
			Moreover, the following inequality holds with probability at least $1-\delta$:
			\begin{align*}
				\big|\widetilde{R}(f)-R(f)\big|
				\le
				&L_\phi C_\ell
				\sqrt{
					\frac{1}{2}\log\frac{2}{\delta}
					\left(
					\frac{m^2}{n_1}+\frac{(m-1)^2}{n_0}
					\right)} \\
				&+
				(\kappa+1)(2m-1)C_\ell \, \Delta_f,
			\end{align*}
			where $L_\phi=\max\{1,\kappa\}$ is the Lipschitz constant of $\phi$.
		\end{theorem}
		
		Theorem~\ref{theorem3} follows the standard bias-and-consistency analysis of corrected risk estimators. For a fixed predictor whose supervised risk is bounded away from zero, the probability that correction is activated decays exponentially with the response-group sample sizes. Hence the correction-induced bias becomes asymptotically negligible, and the corrected estimator remains consistent for that predictor.

		\section{Experiments}\label{sec:experiments}
		This section evaluates the proposed objectives strictly within the synthetic symmetric-query regime studied in the theory. The experiments are designed to address three questions:
		whether direct learning from the original $(L,s)$ supervision is practically feasible, how often the raw groupwise estimator exhibits
		negative empirical risk and poor optimization behavior in finite samples, and whether simple scalar corrections improve robustness in practice. We first describe the synthetic generation protocol and implementation details, then compare URE/NN/ABS with transformed PLL/MCL baselines under different observable-group proportions, and finally examine optimization behavior, batch-size sensitivity, and the larger-label-space stress test.
		
		The experiments are designed to evaluate the statistical mechanism studied in the theory, rather than to simulate all possible human annotation processes. Therefore, we generate query-response triples using the same symmetric random-query mechanism as in Assumption~\ref{assumption1}. This design allows us to isolate whether the proposed risk identity and corrected objectives are effective when the theoretical observation model is satisfied. Extensions to non-uniform, noisy, or adaptive query responses are outside the scope of the present experiments.
		
		\begin{table*}[!t]
			\centering
			\caption{Experimental datasets and statistics on the models used. 5-C and 2-F NNs represent neural networks with 5 convolutional layers and 2 fully connected layers, respectively.}
			\label{tab:accur}
			\begin{tabular}{lccccc}
				\toprule
				\textbf{Dataset} & \textbf{Training} & \textbf{Testing} & \textbf{Features} & \textbf{Classes} & \textbf{Model} \\
				\midrule
				\textbf{MNIST} & 60000 & 10000 & 784 & 10 & CNN \\
				\textbf{FashionMNIST} & 60000 & 10000 & 784 & 10 & CNN\\
				\textbf{KMNIST}  & 60000 & 10000 & 784 & 10 & CNN \\
				\textbf{SVHN} & 50000 & 10000 & 3072 & 10 & ResNet-34 \\
				\textbf{CIFAR-10} & 50000 & 10000 & 3072 & 10 & ResNet-34 \\
				\textbf{USPS} & 5000 & 1000 & 256 & 10 & CNN \\
				\textbf{EMNIST-Letters} & 124800 & 20800 & 784 & 26 & CNN\\
				\bottomrule
			\end{tabular}
		\end{table*}
		\begin{table*}[!t]
			\centering
			\setlength{\tabcolsep}{10pt}
			\renewcommand{\arraystretch}{1.35}
			\caption{Main comparison under the MAE base loss. We report mean test accuracy (\%) $\pm$ standard deviation over five independent runs on six datasets and three values of the observable-group proportion $P(s=1)$. The proposed method is trained with the MAE loss. Trans-MCL (EXP) uses the original EXP upper-bound loss, Trans-MCL (MAE) and Trans-PLL-Avg use the same MAE-type base loss as the proposed method, and Trans-PRODEN is reported under its faithful CCE-based configuration. The best result for each dataset is highlighted in bold.}
			\label{tab:mae}
			
			\begin{tabular}[htbp]{c|l|c c c c c}
				\hline
				$P(s=1)$&Datasets&Ours&Trans-MCL (EXP)&Trans-MCL (MAE)&Trans-PLL-Avg&Trans-PRODEN\\
				\hline
				
				\multirow{6}{*}{0.3}
				& MNIST        & \textbf{98.71 $\pm$ 0.07}   & 98.10 $\pm$ 0.08     &   95.49 $\pm$ 0.05   &  93.80 $\pm$ 4.30  & 98.64 $\pm$ 0.14  \\
				& FashionMNIST & 85.64 $\pm$ 2.44    & 86.34 $\pm$ 0.30     &  82.71 $\pm$ 0.30    &  78.99 $\pm$ 5.43 & \textbf{88.44 $\pm$ 0.36}   \\
				& KMNIST       & \textbf{89.13 $\pm$ 0.42}     & 81.79 $\pm$ 1.68   &  84.64 $\pm$ 0.91    & 69.63 $\pm$ 6.11  & 89.01 $\pm$ 0.25    \\
				& USPS         & \textbf{95.40 $\pm$ 0.16}     & 91.79 $\pm$ 0.18     &  94.44 $\pm$ 0.31    &  87.23 $\pm$ 5.39 & 93.37 $\pm$ 0.39   \\
				& SVHN         & \textbf{94.50 $\pm$ 0.11}    &  92.26 $\pm$ 0.30    &  78.83 $\pm$ 0.64    & 80.62 $\pm$ 1.66  & 92.56 $\pm$ 0.18    \\
				& CIFAR-10     & \textbf{68.90 $\pm$ 3.46}    & 68.39 $\pm$ 0.95   &  60.71 $\pm$ 0.96    & 55.70 $\pm$ 3.09  & 68.60 $\pm$ 0.08   \\
				\cline{1-7}
				
				\multirow{6}{*}{0.5}
				& MNIST        & \textbf{98.86 $\pm$ 0.08}     & 98.27 $\pm$ 0.04     &  98.66 $\pm$ 0.04    & 94.78 $\pm$ 4.92  & 98.73 $\pm$ 0.10   \\
				& FashionMNIST & 83.02 $\pm$ 0.07    &  87.00 $\pm$ 0.26    &  83.32 $\pm$ 0.10    & 79.36 $\pm$ 5.22  & \textbf{88.74 $\pm$ 0.25}   \\
				& KMNIST       & \textbf{89.68 $\pm$ 0.29}    & 84.34 $\pm$ 1.26     &   88.33 $\pm$ 0.29   & 73.10 $\pm$ 4.48  & 89.53 $\pm$ 0.34   \\
				& USPS         & \textbf{95.75 $\pm$ 0.26}    & 92.60 $\pm$ 0.40     & 95.15 $\pm$ 0.33     &  89.88 $\pm$ 0.96 & 93.39 $\pm$ 0.74   \\
				& SVHN         & 89.97 $\pm$ 0.48    & 90.23 $\pm$ 0.13    & 73.30 $\pm$ 1.55     & 80.58 $\pm$ 3.87  & \textbf{91.16 $\pm$ 0.45}   \\
				& CIFAR-10     & 67.77 $\pm$ 4.77     &   \textbf{70.21 $\pm$ 1.02}   & 64.21 $\pm$ 4.52     &  51.27 $\pm$ 1.76  & 70.09 $\pm$ 0.44  \\
				\cline{1-7}
				
				\multirow{6}{*}{0.7}
				& MNIST        &  \textbf{98.73 $\pm$ 0.09}   &  98.12 $\pm$ 0.13    & 95.47 $\pm$ 0.05     &   94.85 $\pm$ 4.96  & 98.62 $\pm$ 0.08  \\
				& FashionMNIST &   84.11 $\pm$ 2.27   & 86.32 $\pm$ 0.14    & 79.91 $\pm$ 5.24     & 82.02 $\pm$ 0.24   & \textbf{88.52 $\pm$ 0.08}  \\
				& KMNIST       &  \textbf{89.26 $\pm$ 0.53}    & 82.35 $\pm$ 1.33     & 86.73 $\pm$ 1.62    &  69.27 $\pm$ 1.65  & 89.22 $\pm$ 0.09  \\
				& USPS         &  \textbf{95.46 $\pm$ 0.46}    & 92.09 $\pm$ 0.13     & 94.53 $\pm$ 0.46     &  88.77 $\pm$ 2.98  & 93.46 $\pm$ 0.63  \\
				& SVHN         &  \textbf{94.48 $\pm$ 0.07}    & 92.07 $\pm$ 0.48     & 66.67 $\pm$ 6.93     & 78.82 $\pm$ 0.91   & 93.02 $\pm$ 0.12    \\
				& CIFAR-10     &  \textbf{70.11 $\pm$ 0.62}    & 70.08 $\pm$ 0.50     &  40.70 $\pm$ 3.26    & 50.52 $\pm$ 1.28   & 69.81 $\pm$ 0.86  \\
				\hline
			\end{tabular}
			
		\end{table*}
		
		\subsection{Synthetic generation for benchmark experiments}
		
		In benchmark experiments with fully labeled datasets, we simulate the observation process in
		Assumption~\ref{assumption1} to construct weakly supervised triples $(x_i,L_i,s_i)$.
		Specifically, for each labeled example $(x_i,y_i)$, we sample $L_i$ uniformly from $\mathcal{Q}_m$ and set
		$s_i=\mathbf{1}\{y_i\in L_i\}$.
		Importantly, the learning algorithm itself does not access $y_i$; it only uses the generated triples $(x_i,L_i,s_i)$.
		Accordingly, the empirical results in this section should be interpreted strictly within the synthetic symmetric-query regime studied in the theory, where the true label is hidden from the learner and only the queried subset together with the binary membership feedback is observed.

		\subsection{Datasets and Experimental Setup}
		We evaluate the method on six standard $k=10$ image classification benchmarks: MNIST\cite{lecun2002gradient}, FashionMNIST\cite{xiao2017fashion}, KMNIST\cite{clanuwat2018deep}, USPS\cite{hull2002database}, SVHN\cite{netzer2011reading}, and CIFAR-10\cite{krizhevsky2009learning}. Table~I summarizes the dataset statistics and model backbones. We additionally report EMNIST-Letters\cite{cohen2017emnist} as a larger-label-space stress test with $k=26$.
		
		For grayscale datasets we use the corresponding CNN backbone, and for color datasets we use ResNet-34. All methods are implemented in PyTorch and trained with Adadelta together with a StepLR scheduler. Unless otherwise stated, we train for 100 epochs with batch size 128. The learning rate is selected from $\{10^{-5},10^{-4},10^{-3},5\times10^{-2},10^{-2},10^{-1},2\times10^{-1}\}$, and the weight decay is selected from $\{10^{-5},10^{-4},10^{-3}\}$. All reported results are means and standard deviations over five independent runs. All experiments are implemented in PyTorch and executed on an NVIDIA GeForce RTX 5090 GPU.

		Unless otherwise stated, for URE/NN/ABS we instantiate the predictor through the softmax output
		\begingroup
		\setlength{\abovedisplayskip}{2pt}
		\setlength{\belowdisplayskip}{2pt}
		\setlength{\abovedisplayshortskip}{2pt}
		\setlength{\belowdisplayshortskip}{2pt}
		\[
		p_\theta(\cdot\mid x)\in\Delta^{k-1},
		\]
		\endgroup
		and use the bounded classwise loss
		\begin{align*}
			&\mathcal{L}_{\mathrm{MSE}}(p_\theta(x),j)=1-2p_\theta(j\mid x)+\sum_{c=1}^{k}p_\theta(c\mid x)^2, \\[2pt]
			&\mathcal{L}_{\mathrm{MAE}}(p_\theta(x),j)=2-2p_\theta(j\mid x), \\[4pt]
			&\mathcal{L}_{\mathrm{GCE}}(p_y; q, \varepsilon) = \frac{1 - \max(p_y, \varepsilon)^q}{q}.
		\end{align*}
		This is the exact loss family used by our implementation for the main URE/NN/ABS experiments. These losses are bounded on the softmax simplex, so the bounded-loss requirement in Section III-D is satisfied by this experimental instantiation.
		
		In the main comparisons, for the transformed MCL baseline, we retain its standard formula and use the EXP upper bound loss and MAE loss.
		
		For each method, the learning rate and weight decay are both selected from the same predefined grids under the same training budget. To keep the comparison controlled, we do not introduce method-specific tuning heuristics beyond these shared search spaces and training horizons, following recent concerns that evaluation protocol and model-selection strategy can materially affect conclusions in weak-label learning \cite{wang2025realistic}. All scalar accuracy values reported in the main comparison tables are summarized as mean $\pm$ standard deviation over five independent runs.
		Table~II reports the main comparison under the MAE base loss, and Table~III reports the comparison of correction methods under the GCE base loss.
		
		\begin{table*}[!t]
			\centering
			\caption{Comparison of correction methods under the GCE base loss. We report mean test accuracy (\%) $\pm$ standard deviation over five independent runs on six datasets and three values of the observable-group proportion $P(s=1)$. The best result for each dataset is highlighted in bold.}
			\label{tbale-cor-gce}
			\setlength{\tabcolsep}{10pt}        
			\renewcommand{\arraystretch}{1.35}  
			\begin{tabular}{c|c|c c c c c c}
				\hline
				$P(s=1)$&method&MNIST&FashionMNIST&KMNIST&USPS&SVHN&CIFAR-10\\
				\hline
				\multirow{3}{*}{0.3}
				& ABS & \textbf{98.66 $\pm$ 0.05} & \textbf{86.44 $\pm$ 0.20} & \textbf{88.18 $\pm$ 0.29} & \textbf{94.90 $\pm$ 0.40} & \textbf{91.32 $\pm$ 0.09} & \textbf{71.87 $\pm$ 0.51} \\
				& NN  & 98.17 $\pm$ 0.12 & 86.16 $\pm$ 0.23 & 85.87 $\pm$ 0.55 & 94.06 $\pm$ 0.34 & 86.68 $\pm$ 0.22 & 70.18 $\pm$ 0.64 \\
				& URE & 95.22 $\pm$ 0.28 & 84.44 $\pm$ 0.48 & 81.42 $\pm$ 0.13 & 89.30 $\pm$ 0.09 & 85.91 $\pm$ 0.19 & 65.44 $\pm$ 0.33 \\
				\cline{1-8}
				\multirow{3}{*}{0.5}
				& ABS  & \textbf{98.63 $\pm$ 0.07} & \textbf{86.47 $\pm$ 0.18} & \textbf{87.44 $\pm$ 0.18} & \textbf{94.19 $\pm$ 0.20} & \textbf{90.58 $\pm$ 0.19} & \textbf{68.74 $\pm$ 0.71} \\
				& NN  & 97.94 $\pm$ 0.05 & 85.93 $\pm$ 0.21 & 84.97 $\pm$ 0.38 & 93.12 $\pm$ 0.26 & 82.72 $\pm$ 0.62 & 64.60 $\pm$ 1.01 \\
				& URE  & 88.81 $\pm$ 0.48 & 79.60 $\pm$ 0.44 & 73.26 $\pm$ 0.70 & 84.36 $\pm$ 0.97 & 70.05 $\pm$ 0.40 & 60.85 $\pm$ 1.33 \\
				
				\cline{1-8}
				\multirow{3}{*}{0.7}
				& ABS & \textbf{98.47 $\pm$ 0.13} & \textbf{84.70 $\pm$ 0.48} & \textbf{84.61 $\pm$ 1.26} & \textbf{93.41 $\pm$ 0.47} & \textbf{88.59 $\pm$ 0.23} & \textbf{60.79 $\pm$ 0.73} \\
				& NN & 97.29 $\pm$ 0.17 & 83.99 $\pm$ 0.34 & 80.71 $\pm$ 0.89 & 91.81 $\pm$ 0.34 & 75.71 $\pm$ 1.40 & 56.32 $\pm$ 0.52 \\
				& URE & 76.92 $\pm$ 0.74 & 68.86 $\pm$ 0.77 & 61.54 $\pm$ 0.91 & 78.15 $\pm$ 0.86 & 40.10 $\pm$ 1.53 & 36.90 $\pm$ 1.61 \\
				\hline
			\end{tabular}
		\end{table*}
		\begin{figure*}[!t]
			\centering
			\captionsetup{font=footnotesize, labelfont=bf, labelsep=colon}
			\subfloat[KMNIST-m3]{\includegraphics[height=3.2cm, width=0.24\textwidth]{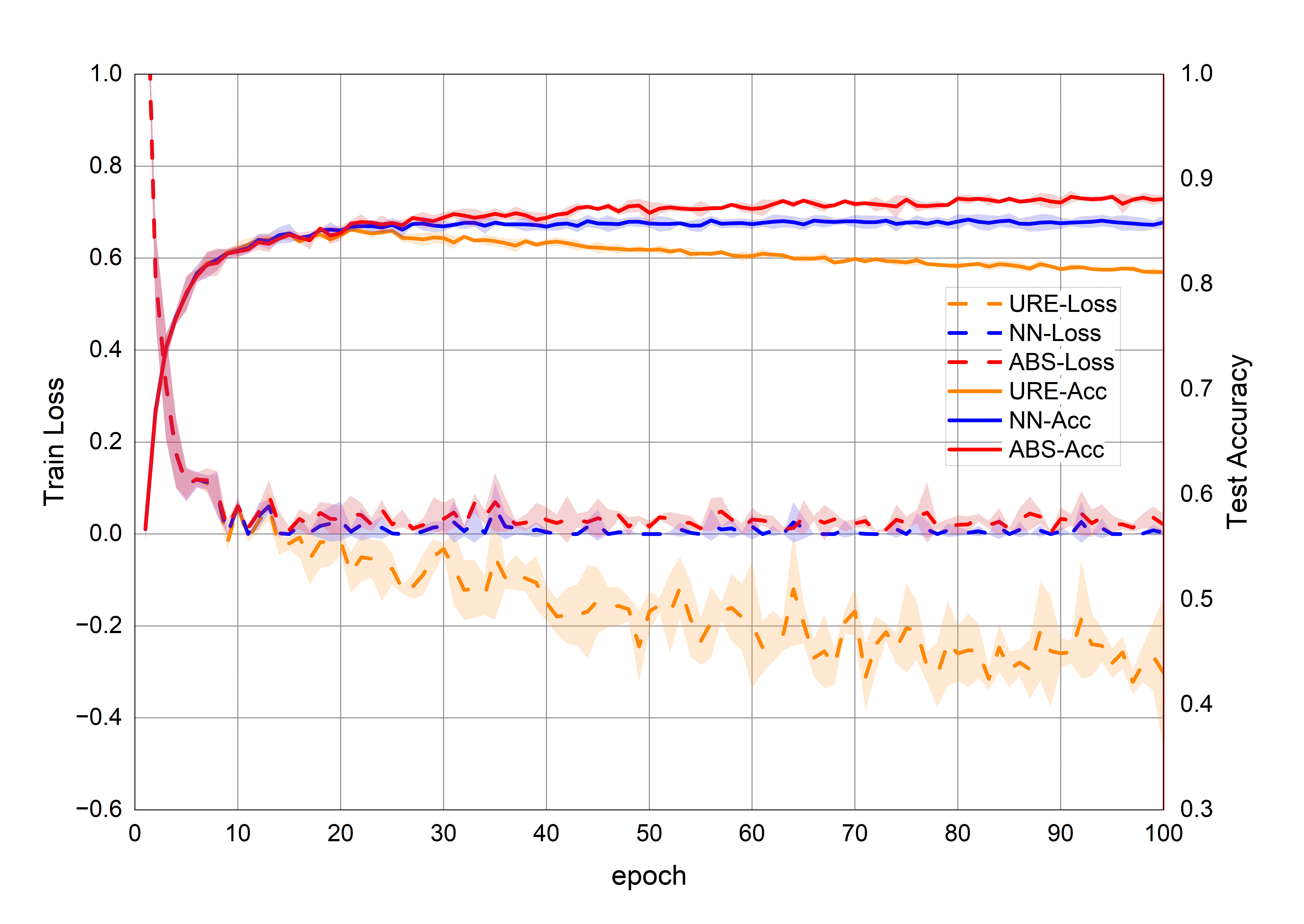}\label{pic-kmnist-m3-gce}}\hfill
			\subfloat[KMNIST-m7]{\includegraphics[height=3.2cm, width=0.24\textwidth]{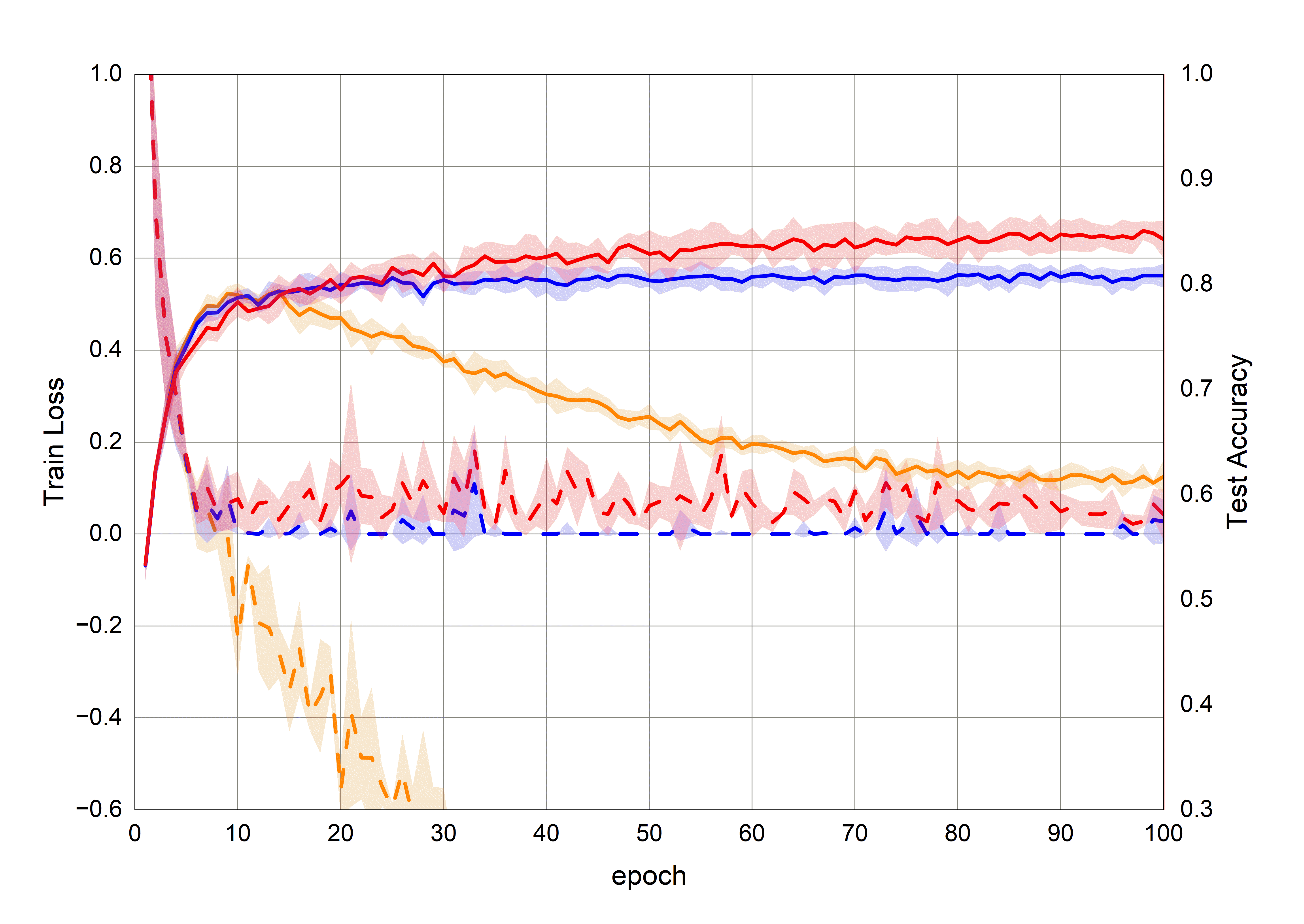}\label{pic-kmnist-m7-gce}}\hfill
			\subfloat[SVHN-m3]{\includegraphics[height=3.2cm, width=0.24\textwidth]{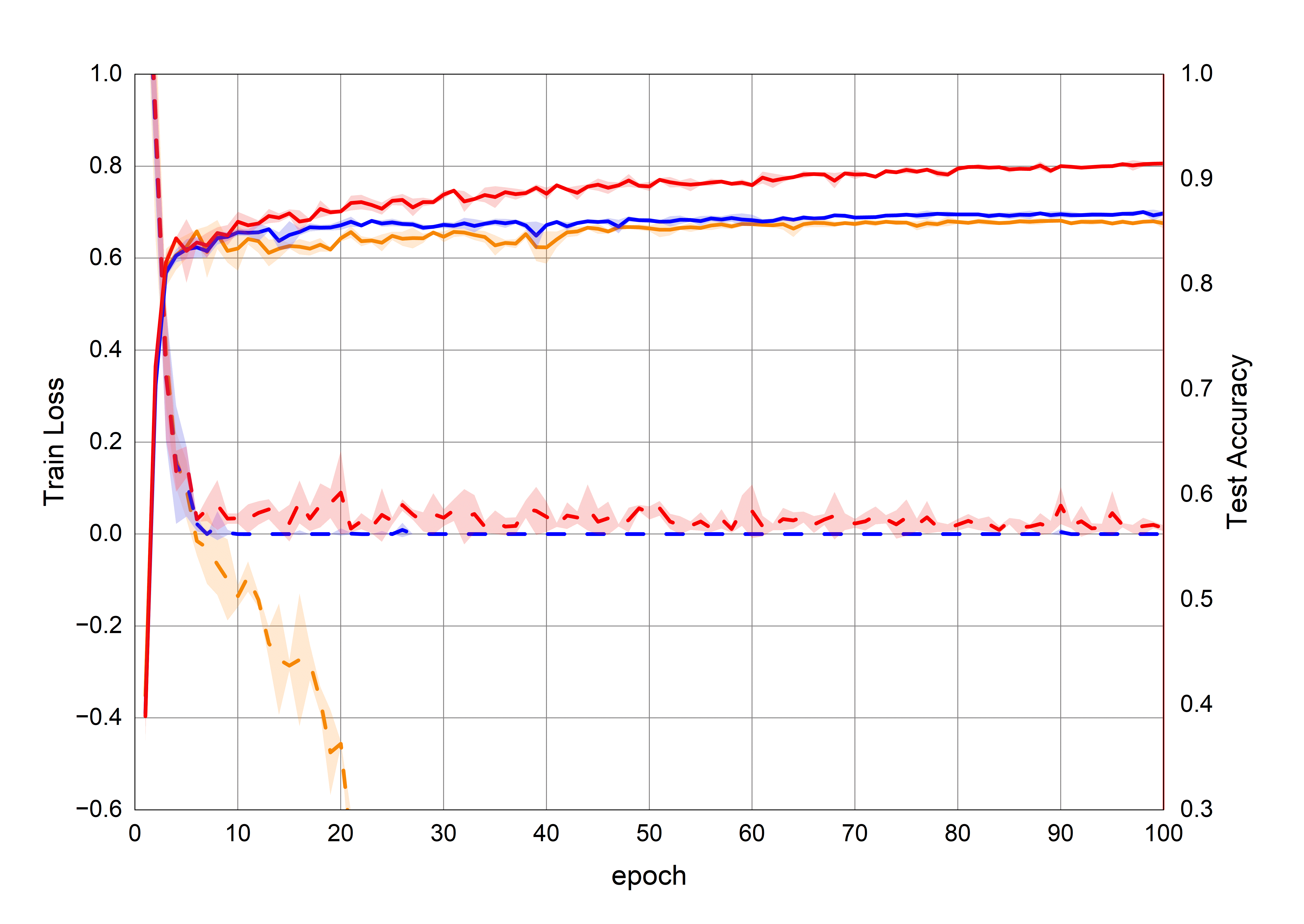}\label{pic-schn-m3-gce}}\hfill
			\subfloat[SVHN-m7]{\includegraphics[height=3.2cm, width=0.24\textwidth]{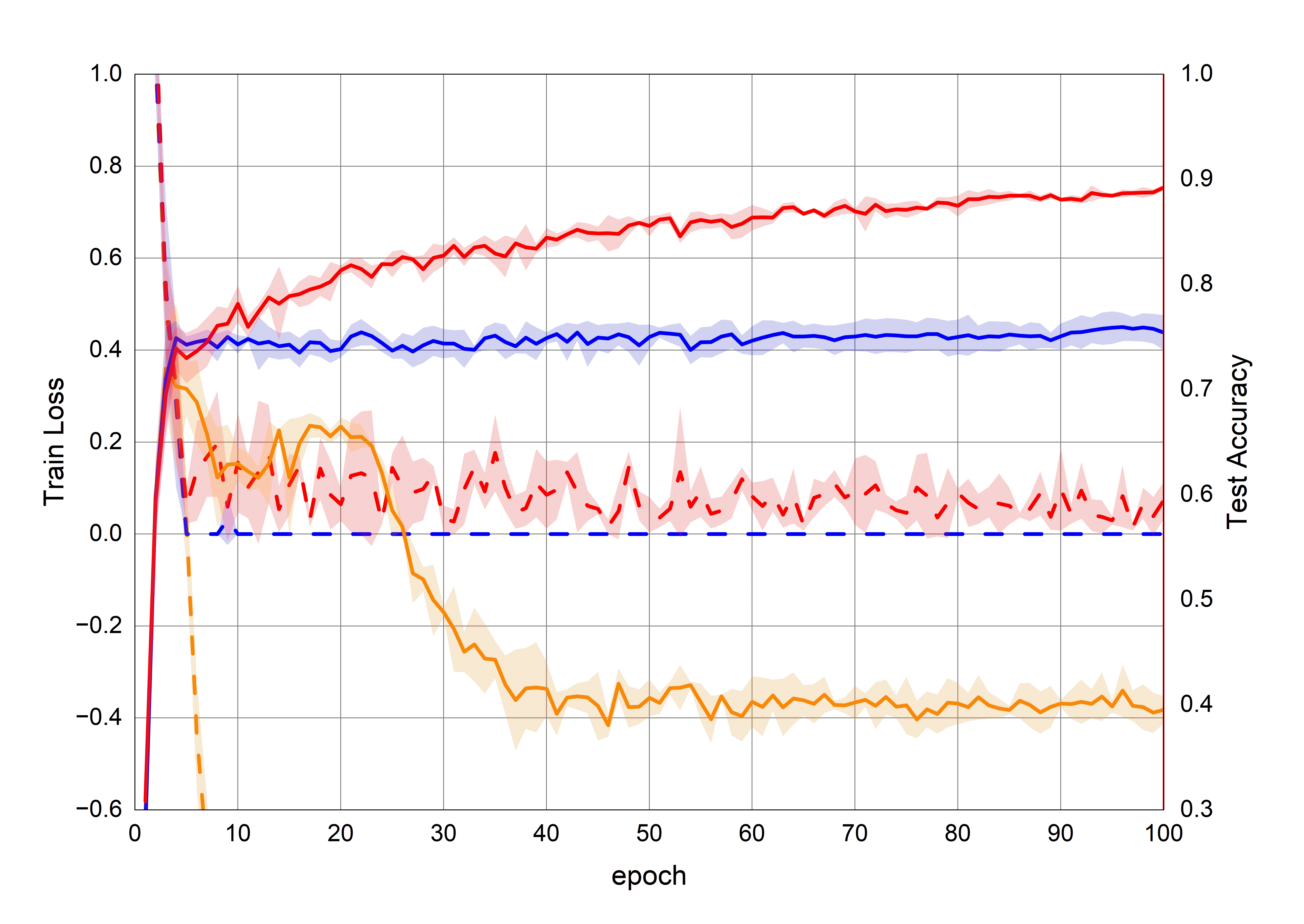}\label{pic-svhn-m7-gce}}\hfill\\[-2pt]
			\caption{Training dynamics under the GCE loss on KMNIST and SVHN. Panels (a) and (c) correspond to $P(s=1)=0.3$ ($m=3$), while panels (b) and (d) correspond to $P(s=1)=0.7$ ($m=7$). Dashed lines denote the training raw/corrected objectives (left axis), and solid lines denote the test accuracy (right axis). Shaded regions indicate mean $\pm$ standard deviation over five independent runs.}
			\label{pic-gce}
		\end{figure*}
		
		\begin{table*}[!t]
			\centering
			\caption{Paired statistical comparison of test accuracy under the GCE loss in challenging settings.}
			\label{tab:paired-gce}
			
			\setlength{\tabcolsep}{14pt}
			\renewcommand{\arraystretch}{1.25}

			\begin{tabular}{c|c|c|c|c}
				\hline
				Dataset & $P(s=1)$ & Comparison & Mean paired diff.(\%) & 95\% CI \\
				\hline
				
				\multirow{2}{*}{KMNIST}
				& \multirow{2}{*}{0.7}
				& NN - URE & 19.17  & [18.96, 19.38]   \\
				& & ABS - URE & 23.07  &  [22.29, 23.85] \\
				\cline{1-5}
				\multirow{2}{*}{SVHN}
				& \multirow{2}{*}{0.7}
				& NN - URE & 35.61 & [33.98, 37.23] \\
				& & ABS -URE & 48.49 & [47.05, 49.92]  \\ 
				\hline
				
			\end{tabular}
		\end{table*}
		
		\subsection{Interpreting the MAE-based main comparison}
		For our softmax-based implementation, the MAE classwise loss is
		\[
		L_{\mathrm{MAE}}(p_\theta(x),j)=2-2p_\theta(j\mid x),
		\qquad p_\theta(\cdot\mid x)\in\Delta^{k-1}.
		\]
		Hence the corresponding supervised target risk is
		\[
		R_{\mathrm{MAE}}(f)
		=
		E_{(x,y)}\!\left[2-2p_\theta(y\mid x)\right]
		=
		2-2\,E_{(x,y)}[p_\theta(y\mid x)],
		\]
		which is manifestly non-negative, since $0\le p_\theta(y\mid x)\le 1$.
		
		For the direct $(L,s)$ objective, define
		\[
		\bar{\ell}_{\mathrm{MAE}}(x,L)
		=
		\frac{1}{m}\sum_{k\in L}\bigl(2-2p_\theta(k\mid x)\bigr)
		=
		2-\frac{2}{m}\sum_{k\in L}p_\theta(k\mid x).
		\]
		Under the symmetric query model, conditioning on $(x,y)$ gives
		\begin{align*}
			&E[\bar{\ell}_{\mathrm{MAE}}(x,L)\mid x,y,s=1] \\
			&= 
			2-\frac{2}{m}\left(
			p_\theta(y\mid x) +\frac{m-1}{k-1}\bigl(1-p_\theta(y\mid x)\bigr)
			\right),
		\end{align*}
		and
		\[
		E[\bar{\ell}_{\mathrm{MAE}}(x,L)\mid x,y,s=0]
		=
		2-\frac{2}{k-1}\bigl(1-p_\theta(y\mid x)\bigr).
		\]
		Therefore,
		\begin{align*}
			&m\,E[\bar{\ell}_{\mathrm{MAE}}\mid x,y,s=1]
			-(m-1)\,E[\bar{\ell}_{\mathrm{MAE}}\mid x,y,s=0] \\
			&=
			2-2p_\theta(y\mid x),
		\end{align*}
		which recovers the same non-negative supervised MAE target at the population level.
		
		We stress, however, that this does not imply that the finite-sample raw empirical objective
		\[
		\widehat R_{\mathrm{raw}}=m\widehat R_1-(m-1)\widehat R_0
		\]
		is guaranteed to remain non-negative for every empirical realization. Rather, it explains why the MAE instantiation is substantially less prone to the pronounced negative-risk instability observed under GCE in our experiments.
		
				\begin{figure*}
			\centering
			\captionsetup{font=footnotesize, labelfont=bf, labelsep=colon}
			\subfloat[KMNIST]{\includegraphics[height=7cm, width=0.5\textwidth]{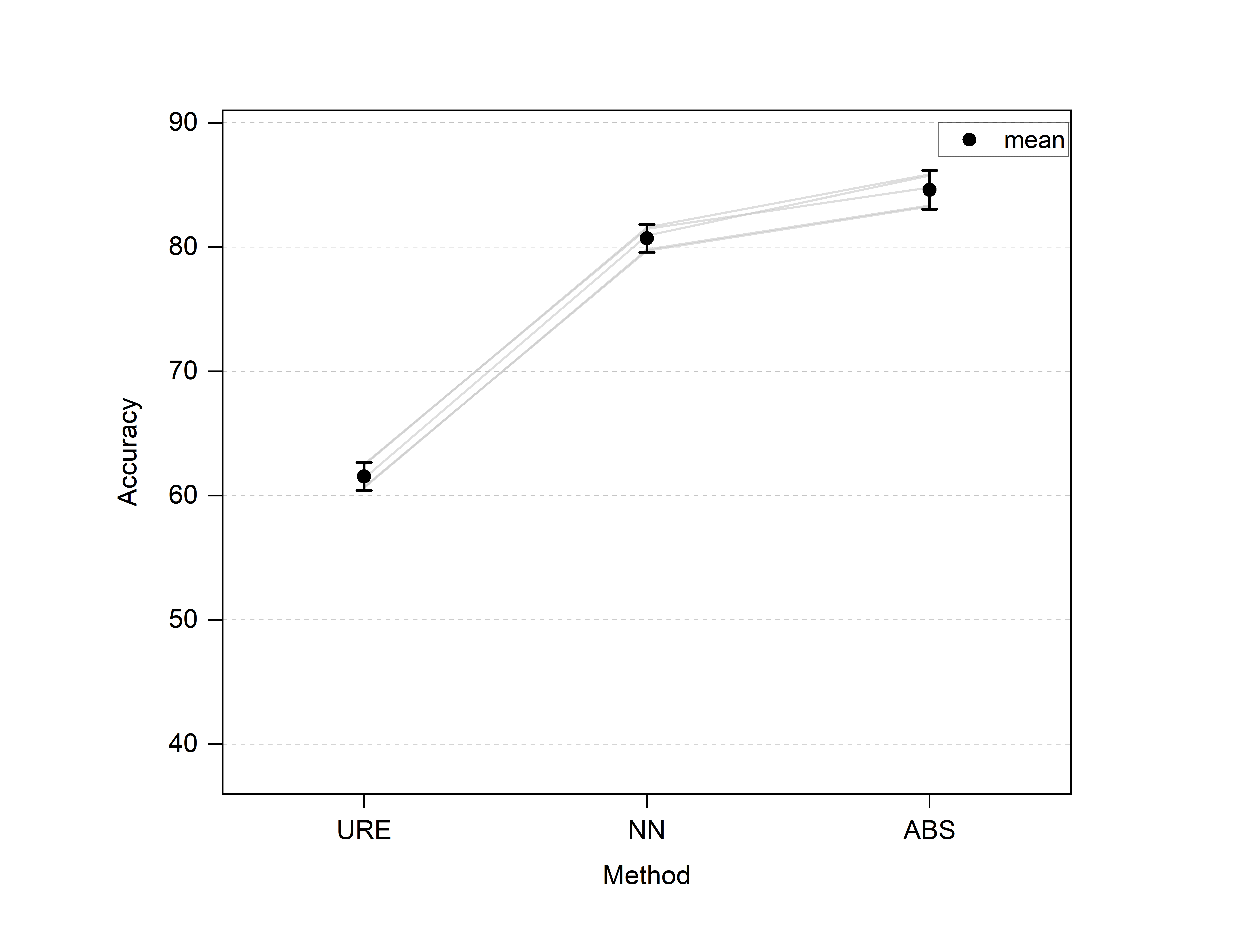}\label{pic-kmnist-m7-err}}  
			\hfill
			\subfloat[SVHN]{\includegraphics[height=7cm, width=0.5\textwidth]{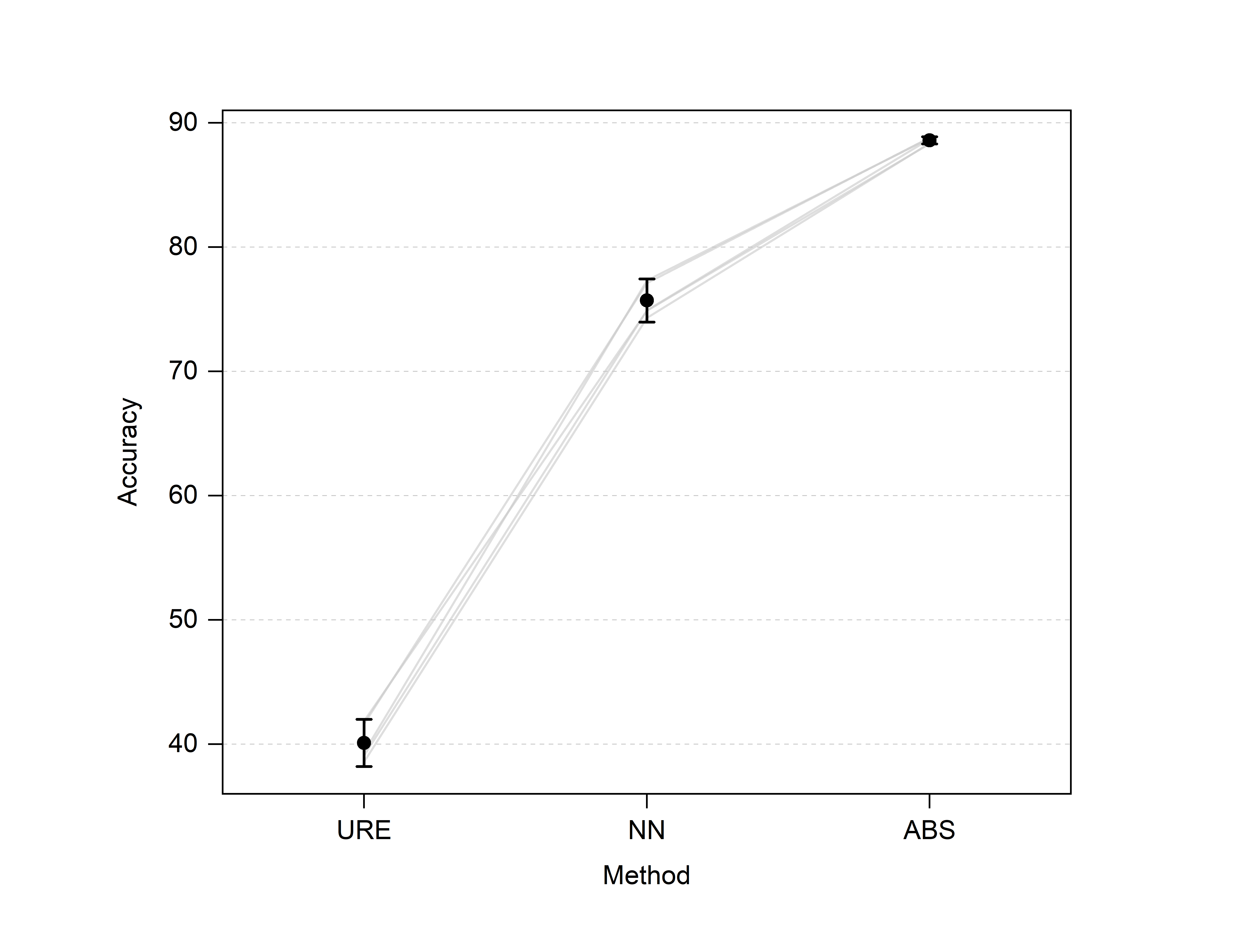}\label{pic-svhn-m7-err}}
			\caption{Paired trial-level comparison of the test accuracy under the GCE loss at $P(s=1)=0.7$. Each thin gray line connects the results of one trial across URE, NN, and ABS, thereby visualizing within-trial performance differences. Black markers denote the mean over five independent runs, and error bars show the 95\% confidence interval. The corrected objectives (NN/ABS) consistently outperform the raw URE objective in these more challenging settings.}
			\label{pic-cor-mean}
		\end{figure*}
		
		\begin{table*}[t]
			\centering
			\caption{Stress test on the EMNIST-Letters dataset ($k=26$), which evaluates a larger label-space regime than the main experiments ($k=10$). We report mean test accuracy (\%) $\pm$ standard deviation over five independent runs under different values of the observable-group proportion $P(s=1)$. The best result is highlighted in bold.}
			\label{tab:emnist_letters}
			
			\begin{threeparttable}
				\setlength{\tabcolsep}{14pt}        
				\renewcommand{\arraystretch}{1.25}  
				
				\begin{tabular}{ccccc}
					\toprule
					$P(s=1)$ & GCE-ABS & GCE-NN & GCE-URE & MAE \\
					\midrule
					0.3 & \textbf{91.16 $\pm$ 0.07} & 90.49 $\pm$ 0.12 & 88.00 $\pm$ 0.21  & 89.15 $\pm$ 1.72  \\
					0.5 & \textbf{90.72 $\pm$ 0.16} & 89.57 $\pm$ 0.14 & 83.05 $\pm$ 0.18 & 72.51 $\pm$ 2.14 \\
					0.7 & \textbf{89.52 $\pm$ 0.32} & 87.45 $\pm$ 0.45 & 71.37 $\pm$ 0.61 & 87.44 $\pm$ 0.16 \\
					\bottomrule
				\end{tabular}
				
				\begin{tablenotes}[flushleft]
					\footnotesize
					\item EMNIST-Letters is reported separately because it changes the experimental regime from $k=10$ to $k=26$.
				\end{tablenotes}
			\end{threeparttable}
		\end{table*}
		
		\begin{table*}[!t]
			\centering
			\caption{Effect of batch size on KMNIST under different values of the observable-group proportion $P(s=1)$. We report mean test accuracy (\%) $\pm$ standard deviation over five independent runs for URE-MAE, GCE-URE, GCE-NN, and GCE-ABS. The best result within each $(P(s=1), \text{batch size})$ block is highlighted in bold.}
			\label{tab:kmnist_batchsize}
			
			\setlength{\tabcolsep}{14pt}
			\renewcommand{\arraystretch}{1.25}

			\begin{tabular}{c|c|c|cccc}
				\hline
				Dataset & $P(s=1)$ & Batch size & MAE & GCE-URE & GCE-NN & GCE-ABS \\
				\hline
				
				\multirow{12}{*}{KMNIST}
				& \multirow{4}{*}{0.3}
				& 128  & \textbf{89.13 $\pm$ 0.42} & 81.42 $\pm$ 0.13  & 85.87 $\pm$ 0.55  & 88.18 $\pm$ 0.29 \\
				& & 256  & \textbf{88.98 $\pm$ 0.33} & 81.22 $\pm$ 0.45  & 85.98 $\pm$ 0.32  & 87.79 $\pm$ 0.39  \\
				& & 512  & \textbf{87.69 $\pm$ 0.24} & 81.32 $\pm$ 0.18 & 85.41 $\pm$ 0.49 & 86.33 $\pm$ 0.33  \\
				& & 1024 & \textbf{85.97 $\pm$ 0.37}  & 82.17 $\pm$ 0.48 & 84.50 $\pm$ 0.43  & 85.01 $\pm$ 0.43 \\
				\cline{2-7}
				
				& \multirow{4}{*}{0.5}
				& 128  & \textbf{89.68 $\pm$ 0.29} & 73.26 $\pm$ 0.70 & 84.97 $\pm$ 0.38 & 87.44 $\pm$ 0.18 \\
				& & 256  & \textbf{88.72 $\pm$ 1.43} & 72.35 $\pm$ 0.22 & 84.65 $\pm$ 0.99 & 86.64 $\pm$ 1.29 \\
				& & 512  & \textbf{87.33 $\pm$ 1.33} & 75.48 $\pm$ 0.26  & 84.06 $\pm$ 0.88 &  85.33 $\pm$ 0.81  \\
				&      & 1024 & \textbf{85.61 $\pm$ 2.40} & 78.46 $\pm$ 1.00  & 83.54 $\pm$ 0.98 & 83.78 $\pm$ 1.12 \\
				\cline{2-7}
				
				& \multirow{4}{*}{0.7}
				& 128  & \textbf{89.26 $\pm$ 0.53}  & 61.54 $\pm$ 0.91 & 80.71 $\pm$ 0.89 & 84.61 $\pm$ 1.26 \\
				&  & 256  & \textbf{88.54 $\pm$ 1.19} & 62.21 $\pm$ 0.58  & 80.85 $\pm$ 0.54  & 83.62 $\pm$ 0.24 \\
				&  & 512  & \textbf{87.79 $\pm$ 0.87} & 62.96 $\pm$ 0.95 & 79.99 $\pm$ 0.47 & 81.28 $\pm$ 0.77 \\
				&  & 1024 & \textbf{86.57 $\pm$ 0.91}  & 69.17 $\pm$ 0.92  & 79.05 $\pm$ 0.58  & 79.73 $\pm$ 0.61  \\
				\hline
				
			\end{tabular}
		\end{table*}
		
		\begin{figure}
			\centering
			\captionsetup{font=footnotesize, labelfont=bf, labelsep=colon}
			\subfloat[EMNIST-Letters]{\includegraphics[height=10cm, width=0.45\textwidth]{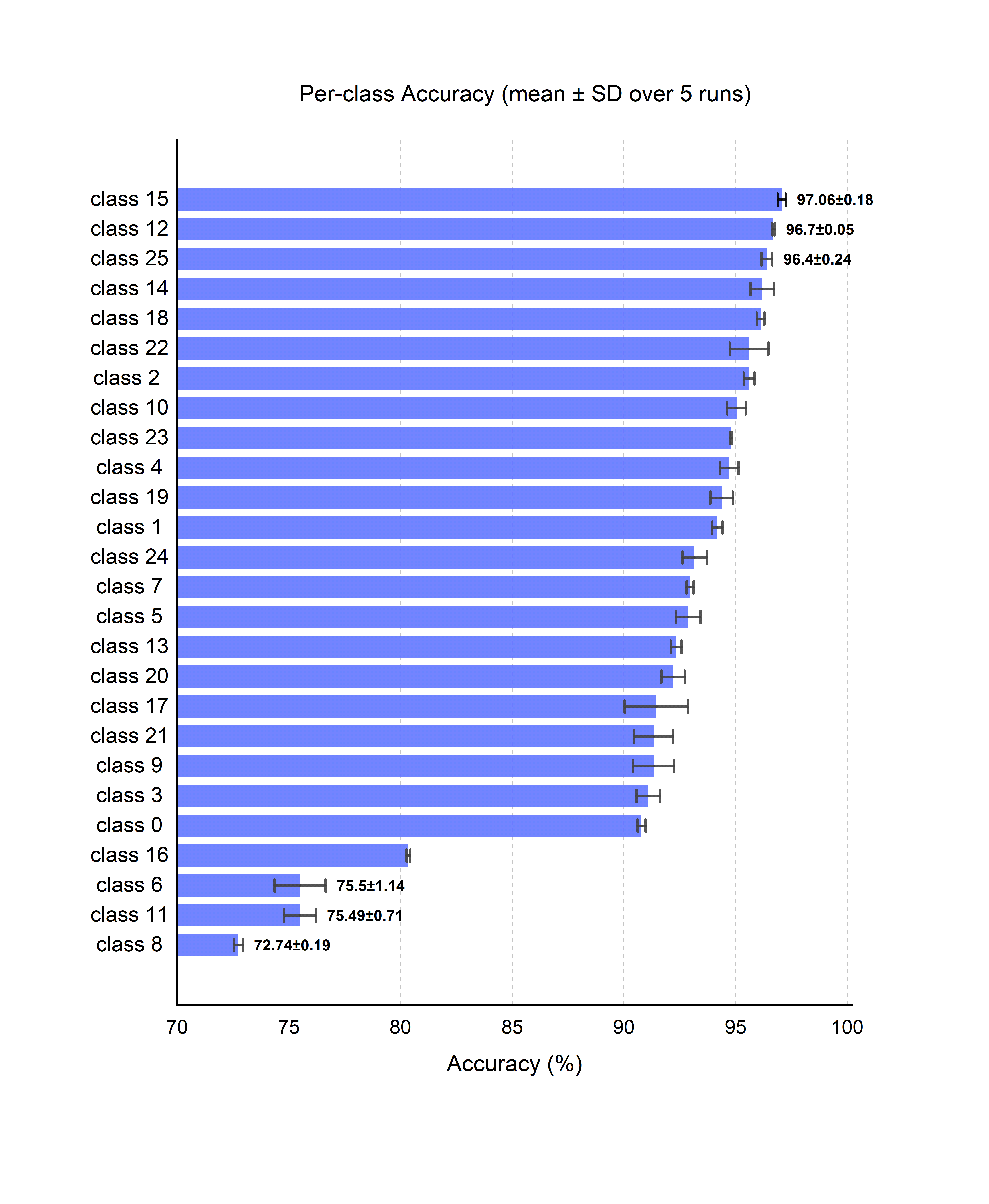}\label{pic:emnist}  }
			\caption{Per-class accuracy on the EMNIST-Letters dataset under GCE-ABS with $P(s=1)=0.3$. Bars show the mean per-class true-positive accuracy over five independent runs, and error bars indicate the standard deviation.} 
			\label{pic-emnist-tpr}
		\end{figure}
		
		\begin{figure*}[htbp]
			\centering
			\captionsetup{font=footnotesize, labelfont=bf, labelsep=colon}
			\subfloat[KMNIST-m3]{\includegraphics[height=4.2cm, width=0.3\textwidth]{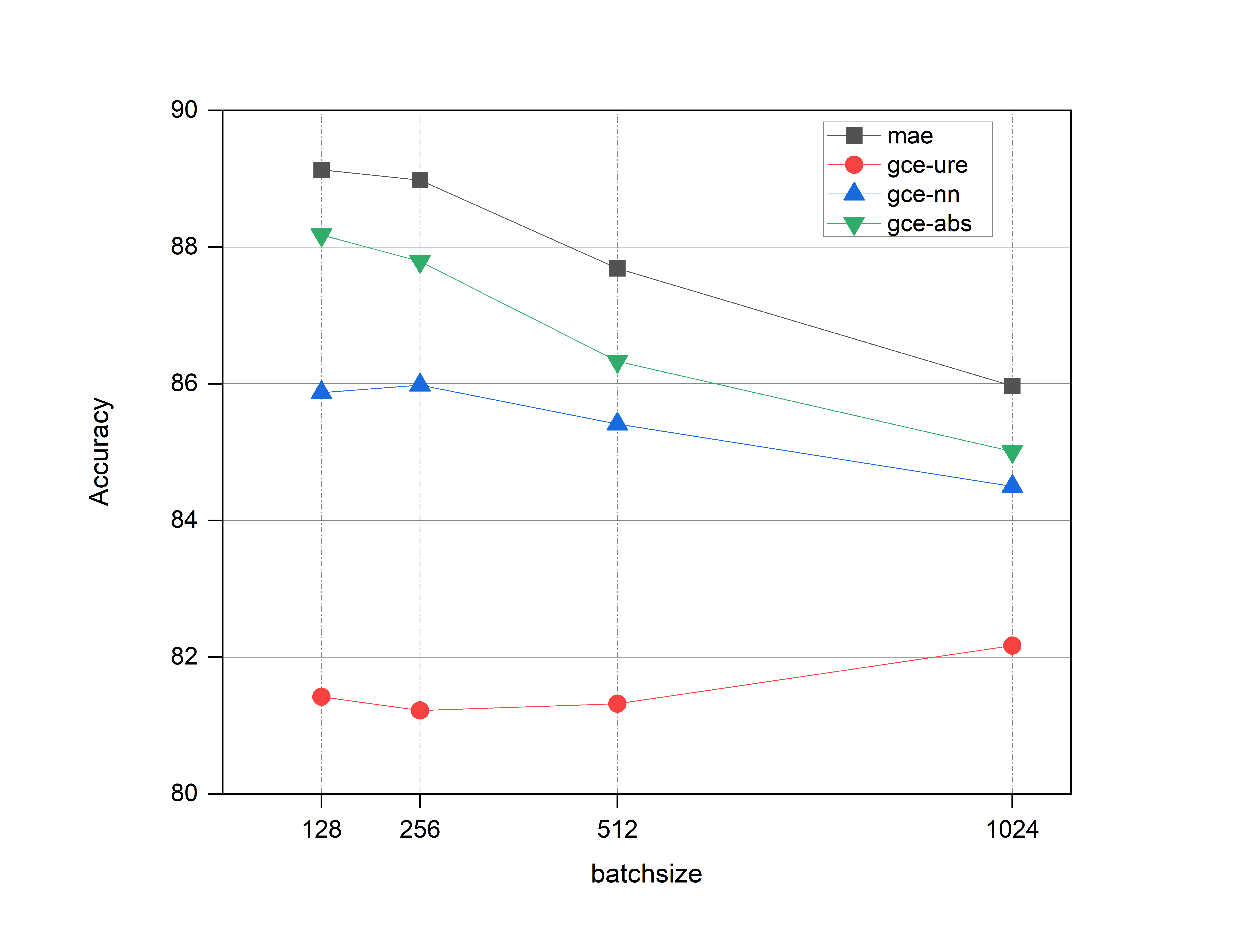}\label{pic-m3-batch}}\hfill
			\subfloat[KMNIST-m5]{\includegraphics[height=4.2cm, width=0.3\textwidth]{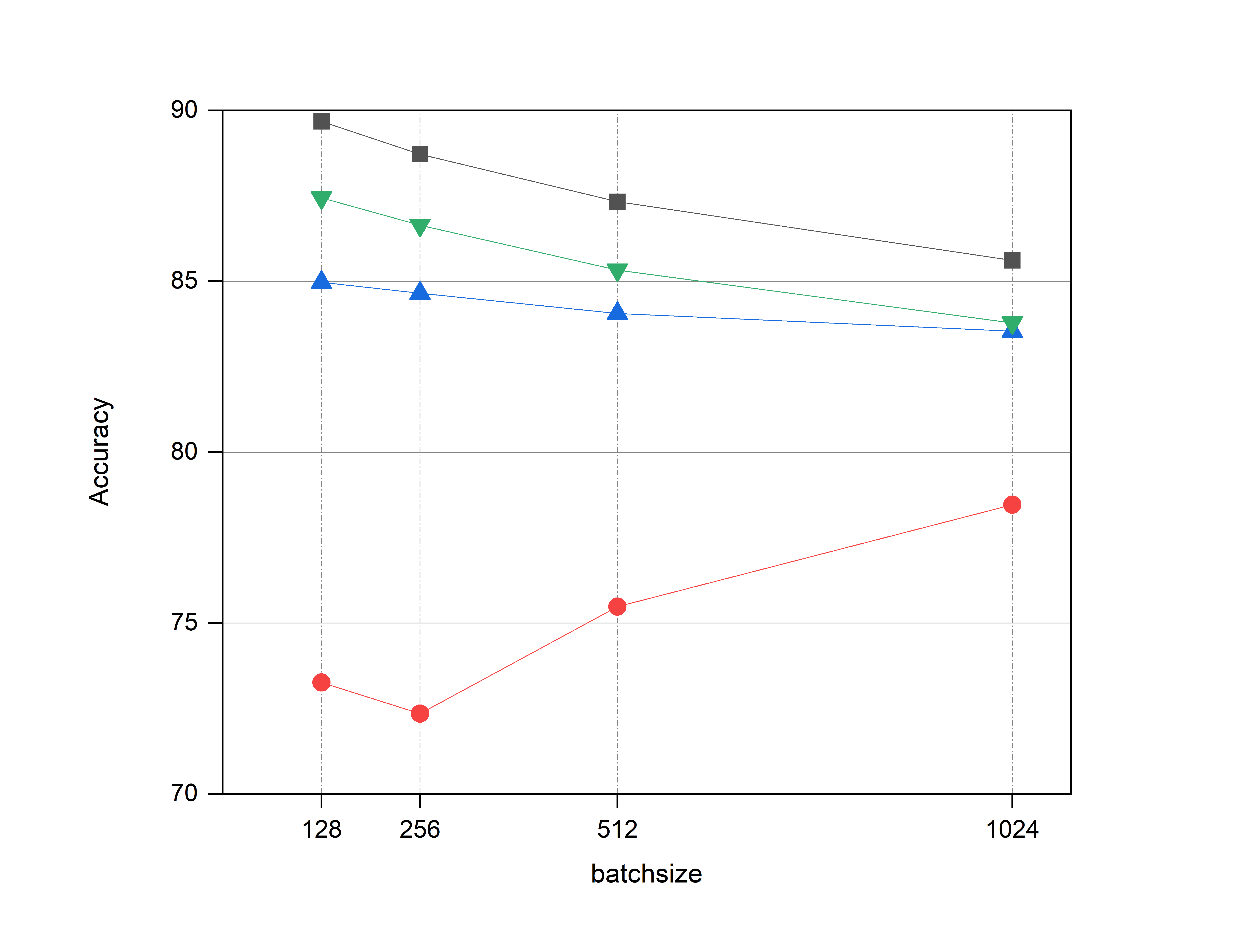}\label{pic-m5-batch}}\hfill
			\subfloat[KMNIST-m7]{\includegraphics[height=4.2cm, width=0.3\textwidth]{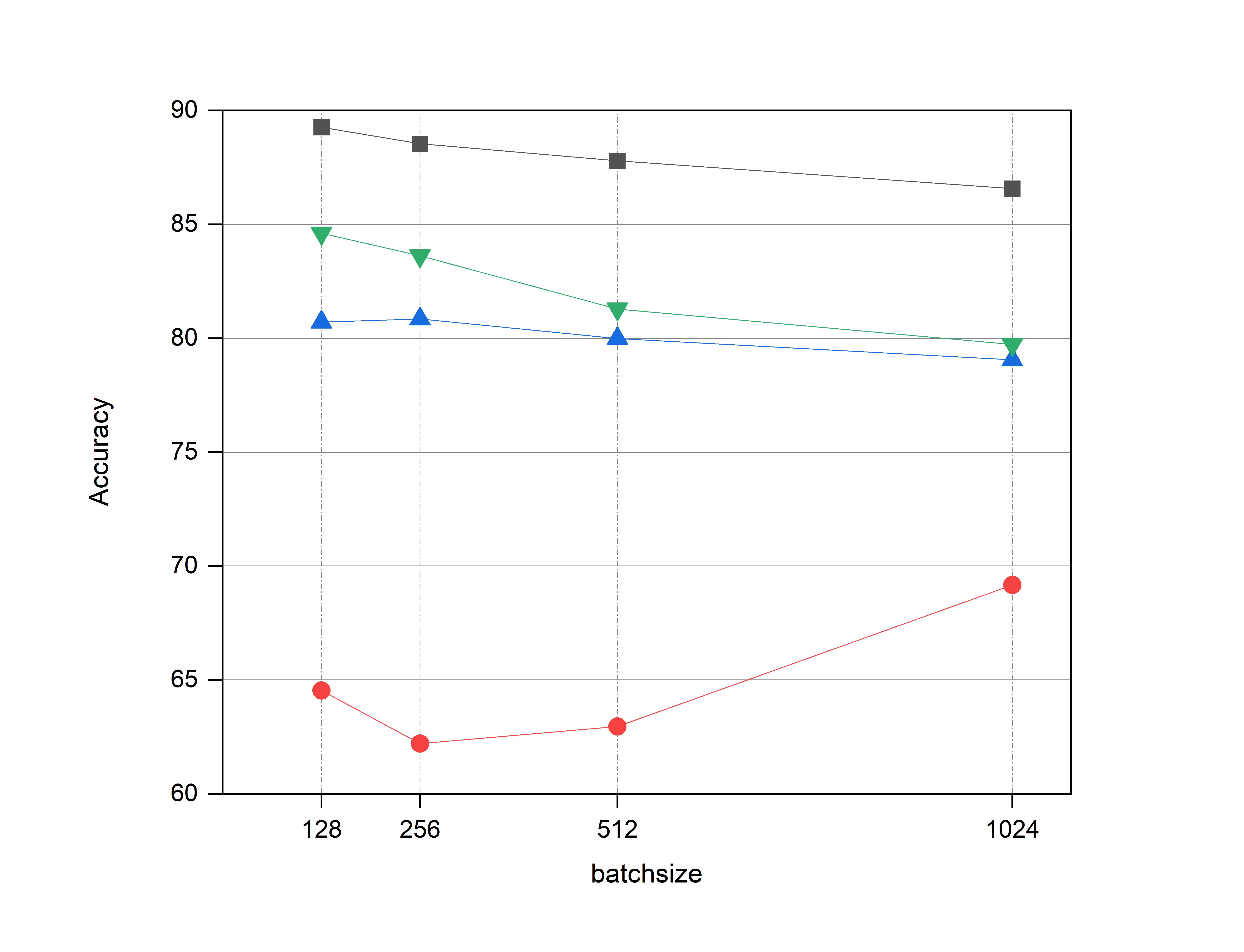}\label{pic-m7-batch}}\hfill\\[-2pt]
			\caption{Sensitivity to batch size on KMNIST under different values of the observable-group proportion $P(s=1)$. Each panel reports the test accuracy averaged over five independent runs for URE-MAE, GCE-URE, GCE-NN, and GCE-ABS. The raw GCE-URE objective is substantially more sensitive to batch size, especially when $P(s=1)$ is large, whereas the corrected objectives are comparatively more stable.}
			\label{pic-batchsize}
		\end{figure*}
		\begin{figure*}[htbp]
			\centering
			\captionsetup{font=footnotesize, labelfont=bf, labelsep=colon}
			\subfloat[Batch size 128]{\includegraphics[height=3.2cm, width=0.24\textwidth]{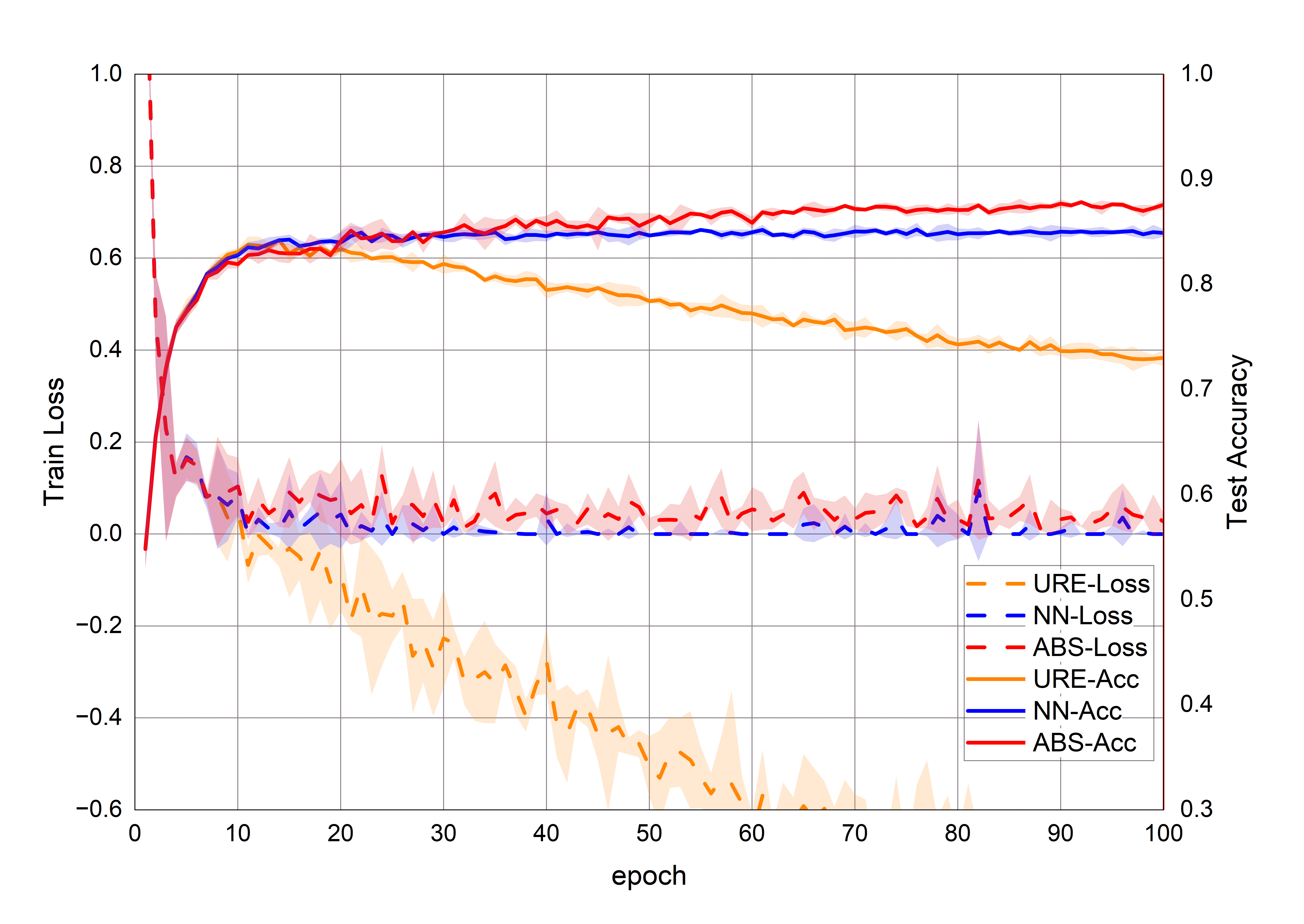}\label{fig:batch128}}\hfill
			\subfloat[Batch size 256]{\includegraphics[height=3.2cm, width=0.24\textwidth]{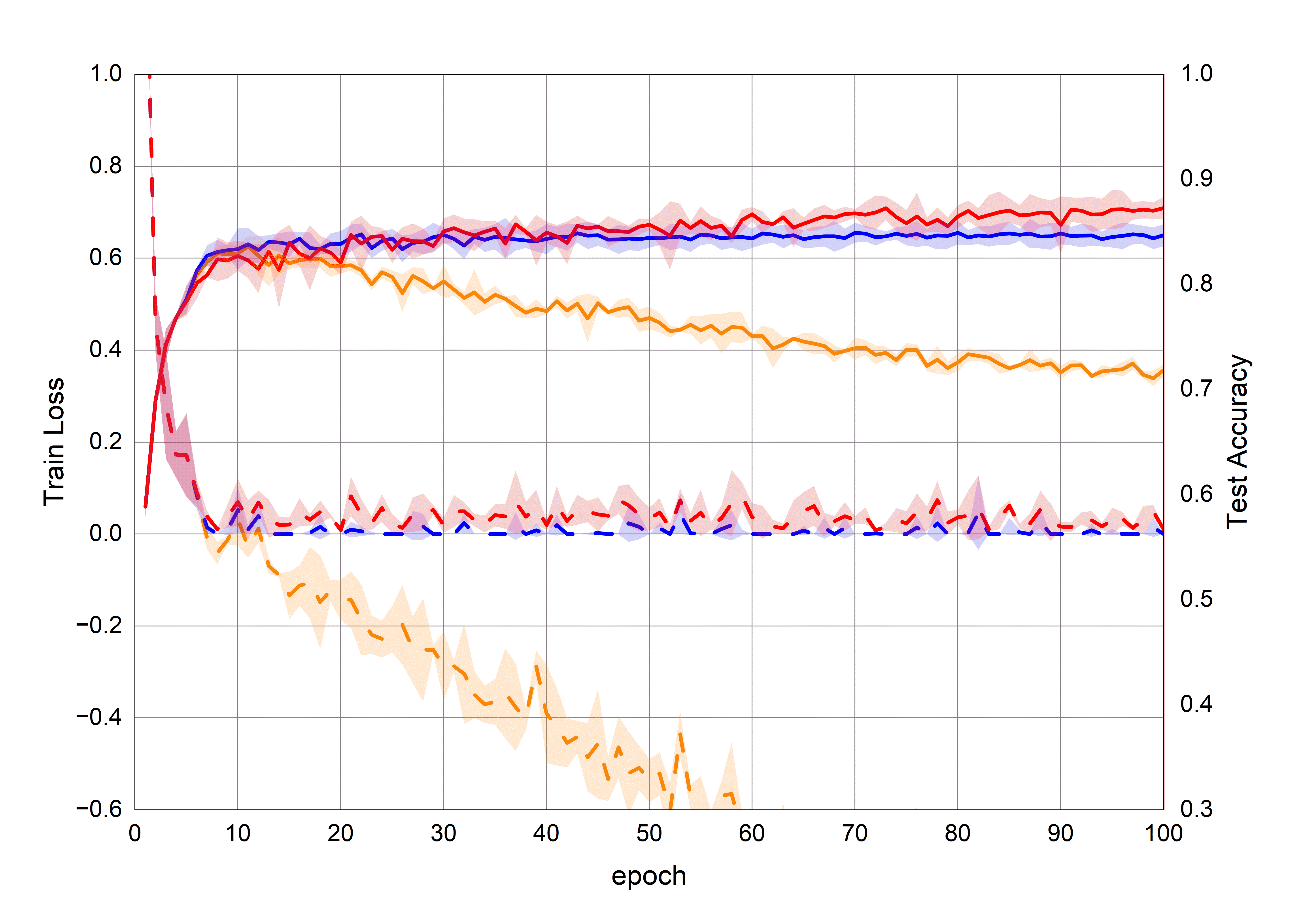}\label{fig:batch256}}\hfill
			\subfloat[Batch size 512]{\includegraphics[height=3.2cm, width=0.24\textwidth]{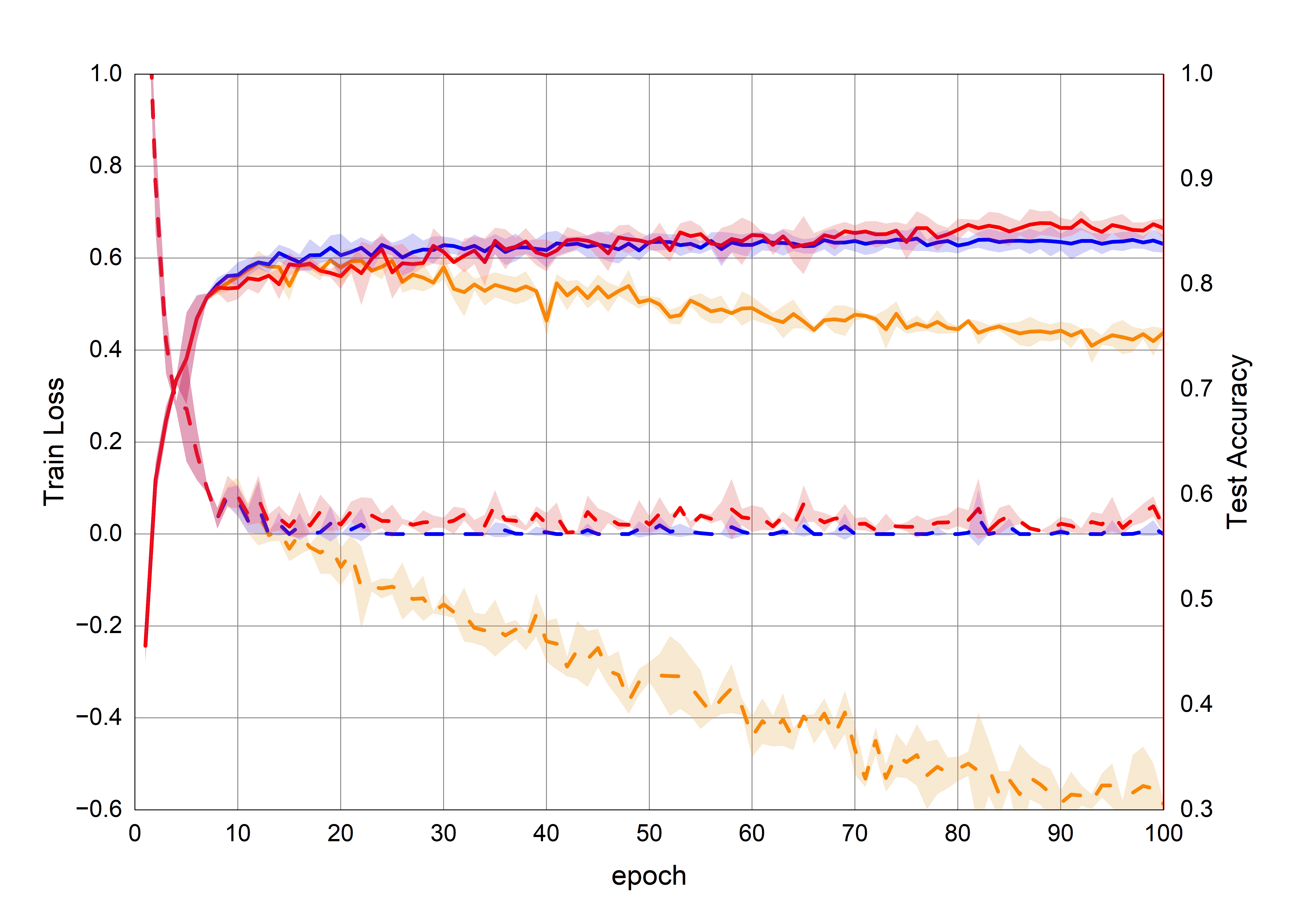}\label{fig:batch512}}\hfill
			\subfloat[Batch size 1024]{\includegraphics[height=3.2cm, width=0.24\textwidth]{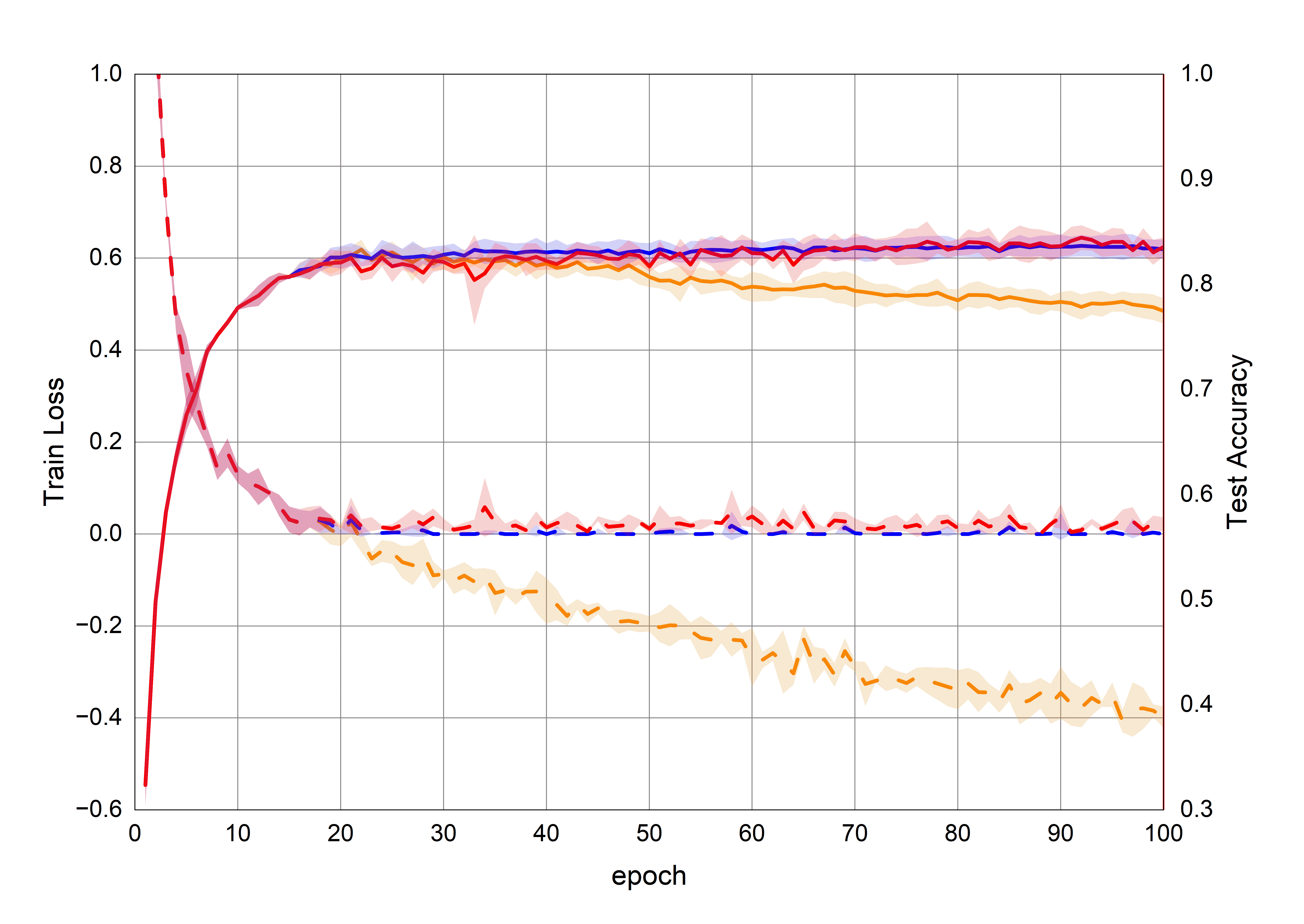}\label{fig:batch1024}}\hfill\\[-2pt]
			\caption{Training dynamics on KMNIST under the GCE loss with $P(s=1)=0.7$ for different batch sizes. Dashed lines denote the training raw/corrected objectives (left axis), and solid lines denote the test accuracy (right axis). Shaded regions indicate mean $\pm$ standard deviation over five independent runs.}
			\label{pic-batch}
		\end{figure*}
		
		\subsection{Baseline Methods}
		
		Our training data consist of triples $(x_i,L_i,s_i)$ generated by randomized subset queries with binary membership feedback. This observation protocol is not the native input format assumed by standard multiple complementary-label learning (MCL) or partial-label learning (PLL) methods. To enable comparison, we first transform each observed pair $(L_i,s_i)$ into the weak-label format required by the corresponding baseline, and then apply the original baseline method on the transformed data.
		
		We include four transformed baselines. First, we apply the original MCL upper-bound method with its EXP loss, denoted \textbf{TMCL (EXP)}; this baseline follows the original MCL formulation after converting each query-response observation into a valid complementary-label set \cite{feng2020learning}. Second, we report a controlled MCL comparison under the same MAE-type base loss as our direct objective, denoted \textbf{TMCL (MAE)} \cite{feng2020learning}. Third, we include an average-based PLL baseline, denoted \textbf{Trans-PLL-Avg}, after converting each observation into a valid candidate-label set \cite{cour2011learning}. Finally, we include \textbf{Trans-PRODEN}, which applies the original PRODEN method to the same transformed PLL-style candidate sets and keeps its standard confidence-update mechanism and cross-entropy-based configuration \cite{lv2020progressive}.
		
		All transformed baselines use the same backbone, optimizer family, epoch budget, and learning-rate decay schedule as the proposed method. We distinguish between \emph{faithful} comparisons, which preserve the original loss design of the baseline, and \emph{controlled} comparisons, which align the base loss whenever such alignment is meaningful. The transformed baselines follow the conversion rules described above and use their original objectives whenever applicable.
		
		\subsection{Results and interpretation}
		
		The experiments mirror the logic of Sections III and IV: we first test whether the direct query-response objective is viable under the matched mechanism, and then examine whether correction improves optimization when the raw weighted-difference objective becomes unstable.
		
		We use MAE as the base loss in the main comparison because, under the direct $(L,s)$ objective, it is empirically much more stable than GCE. This does not remove the structural issue identified in Section~IV: training still optimizes a difference of two sample means, and the raw empirical objective may therefore take unfavorable values in finite samples. In the present experiments, the MAE instantiation is substantially less prone to the pronounced negative-risk behavior that frequently appears under GCE. For this reason, Table~II uses MAE for the main bounded-loss comparison, whereas Table~III and Figs.~\ref{pic-gce}--\ref{pic-cor-mean} use GCE to highlight the empirical stabilization effect of the corrected objectives.
		
		Table~II reports the main comparison under the MAE base loss. The main conclusion is feasibility rather than dominance. Under the matched synthetic query mechanism, direct learning from the original $(L,s)$ supervision is empirically viable and often competitive with transformed weak-label baselines. The proposed method is consistently strong on MNIST, KMNIST, and USPS, remains competitive on SVHN and CIFAR-10, and is not uniformly best on FashionMNIST. This is the appropriate scope of interpretation. Table~II supports the claim that the native hit/miss objective provides a viable mechanism-specific training route in the controlled regime studied here, but it does not support a blanket superiority claim over all transformed baselines in all settings.
		
		Table~III isolates the role of correction under the GCE loss, where the raw estimator is much more sensitive to finite-sample fluctuation. Here the pattern is considerably sharper than in Table~II. ABS is best across all six datasets and all three values of the observable-group proportion, NN consistently improves over the raw URE objective, and the uncorrected objective deteriorates substantially as $P(s=1)$ increases. This behavior is closely aligned with the theory in Section~IV: once the raw direct estimator becomes more instability-prone, direct optimization of the uncorrected difference-of-means objective becomes unreliable, whereas simple scalar corrections substantially improve robustness.
		
		The qualitative training dynamics in Fig.~\ref{pic-gce} are consistent with the quantitative comparison in Table~III. In the settings, such as KMNIST-$m3$ and SVHN-$m3$, all three objectives eventually learn useful predictors, although the
		corrected objectives are empirically more stable. In the harder settings, especially KMNIST-$m7$ and SVHN-$m7$, the raw URE objective frequently drops into a strongly negative regime and its test accuracy plateaus at a much lower level. By contrast, NN and ABS remain more stable in practice and achieve
		clearly better test accuracy. Table~IV and Fig.~\ref{pic-cor-mean} provide paired trial-level evidence for this effect at $P(s=1)=0.7$. On KMNIST, the mean paired improvements over URE are $19.17\%$ for NN and $23.07\%$ for ABS, with 95\% confidence intervals $[18.96,19.38]$ and $[22.29,23.85]$. On SVHN, the corresponding improvements are $35.61\%$ and $48.49\%$, with 95\% confidence intervals $[33.98,37.23]$ and $[47.05,49.92]$. Since all reported intervals lie strictly above zero, the paired comparison supports the visual evidence in Fig.~\ref{pic-cor-mean} that the corrected objectives consistently improve over the raw URE objective in these challenging GCE settings.
		
		Table~V extends the evaluation to EMNIST-Letters, where the label space increases from $k=10$ to $k=26$. This experiment is not intended as a realistic large-scale benchmark, but rather as a controlled stress test for the dependence on label-space size. The qualitative pattern remains similar to the $k=10$ experiments: GCE-ABS is best across all three values of the observable-group proportion, GCE-NN is second, and GCE-URE degrades as $P(s=1)$ increases. In particular, GCE-URE drops from $88.00\%$ at $P(s=1)=0.3$ to $71.37\%$ at $P(s=1)=0.7$, whereas GCE-ABS remains above $89\%$ throughout. URE-MAE is more stable than GCE-URE, but still remains below GCE-ABS. Fig~\ref{pic-emnist-tpr} further shows that under GCE-ABS with $P(s=1)=0.3$, most classes achieve high true-positive accuracy, while only a few difficult classes account for most of the remaining error. Taken together, Table~V and Fig.~\ref{pic-emnist-tpr} suggest that the correction-based advantages are not restricted to the $k=10$ regime.
		
		Table~VI and Fig.~\ref{pic-batchsize} examine sensitivity to batch size on KMNIST. The main pattern is not that larger batches uniformly improve all methods, but rather that the raw GCE-URE objective is much more sensitive to batch size than the corrected objectives. At $P(s=1)=0.7$, GCE-URE improves from $61.54\%$ with batch size $128$ to $69.17\%$ with batch size $1024$, which is consistent with reduced fluctuation of the raw
		difference-of-means estimator at larger batch sizes. Nevertheless, even at the largest batch size, GCE-URE remains clearly below GCE-NN and GCE-ABS. By contrast, the corrected GCE objectives vary more moderately across batch sizes and preserve a substantial advantage over raw URE throughout. URE-MAE is comparatively stable, but its performance still decreases somewhat as the batch size grows. Fig~\ref{pic-batch} provides the corresponding optimization trajectories for $p(S = 1)=0.7$: smaller batches make the raw GCE-URE objective more erratic in practice and more prone to negative values, whereas the corrected objectives remain much better behaved empirically.
		
		Overall, the experiments support a focused empirical conclusion that is fully consistent with the preceding theory. Under the synthetic symmetric-query mechanism studied in this paper, direct learning from the original query-response supervision is practically feasible. When the base loss yields a comparatively stable direct objective, as in the MAE comparison, the direct route is already competitive. When the raw objective becomes more instability-prone, as under GCE, the correction-based variants become practically important and markedly more reliable. The empirical evidence therefore supports direct query-response learning and risk correction under the matched query mechanism, rather than a broad claim of universal superiority over transformed weak-label alternatives.
		
		\section{Conclusion}
		
		This paper studied multiclass learning from random label-subset membership queries, where each instance is observed through a queried label subset and a binary membership response. Under a symmetric random-query mechanism, we characterized the induced query-response distribution and derived an exact rewriting of the ordinary supervised multiclass risk. The resulting unbiased empirical risk estimator has a weighted difference-of-means structure over two response groups.
		
		This structure is central to the proposed framework. It enables direct risk estimation from query-response observations, but it also explains why the empirical risk can go negative in finite samples. To address negative empirical risk and the associated overfitting problem, we introduced corrected risk estimators based on non-negative and absolute-value corrections. For the raw unbiased estimator, we established conditional generalization and excess-risk guarantees with explicit dependence on the two response-group sample sizes. For the corrected estimator, we established a bias-and-consistency result under a positive-risk condition for a fixed predictor.
		
		Experiments under the matched synthetic query mechanism show that direct query-response learning is feasible and that risk correction improves empirical stability when the raw objective becomes unstable. These results suggest that, when supervision is collected through structured subset questions, the query mechanism itself should be taken into account in risk estimation and objective design. Extending the analysis to non-uniform query distributions, noisy responses, adaptive query design, and real annotation protocols remains an important direction for future work.

		\bibliographystyle{ieeetr}
		\bibliography{ref}

	\end{document}